\documentclass[10pt,journal,compsoc]{IEEEtran}

\usepackage{color}
\usepackage{caption}
\usepackage{times}
\usepackage{helvet}
\usepackage{courier}
\usepackage{amsmath}
\usepackage{amssymb}
\usepackage{amsthm}
\usepackage{graphicx}
\usepackage{array}
\usepackage{booktabs}
\usepackage{amsopn}
\usepackage{amsmath,bm}
\usepackage{algorithm}
\usepackage{algorithmic}
\usepackage{enumerate}
\usepackage{color}
\usepackage{multirow}
\usepackage{bbding}
\usepackage{threeparttable}
\usepackage{threeparttable} 
\usepackage{dsfont} 
\usepackage{makecell}
\usepackage{url}  
\usepackage{caption}
\usepackage{subcaption}
\usepackage[numbers]{natbib}
\usepackage{enumitem}
\usepackage{mathtools}
\usepackage{tcolorbox}
\usepackage{hyperref}

\newcommand{\eg}{\textit{e.g., }}
\newcommand{\etal}{\emph{et al.}}
\newcommand{\ie}{\emph{i.e.}}
\newcommand{\etc}{\emph{etc}}

\graphicspath{{figs/}}
\begin{document}

\title{A Systematic Survey of Prompt Engineering on Vision-Language Foundation Models}

\author{Jindong Gu,\; Zhen Han,\; Shuo Chen,\; Ahmad Beirami,\; Bailan He,\; Gengyuan Zhang,\; Ruotong Liao,\; \\ Yao Qin,\; Volker Tresp,\; Philip Torr
\IEEEcompsocitemizethanks{
\IEEEcompsocthanksitem Jindong Gu and Philip Torr are with University of Oxford.
\IEEEcompsocthanksitem Zhen Han, Shuo Chen, Bailan He are with Ludwig Maximilian University of Munich.
\IEEEcompsocthanksitem Ahmad Beirami is with Google Research.
\IEEEcompsocthanksitem Gengyuan Zhang, Ruotong Liao, and Volker Tresp are with Ludwig Maximilian University of Munich and Munich Center for Machine Learning.
\IEEEcompsocthanksitem Yao Qin is with Google Research and University of California, Santa Barbara.
\IEEEcompsocthanksitem Corresponding E-mail from Jindong Gu: \{jindong.gu\}@outlook.com.
}
}

\IEEEtitleabstractindextext{%
\begin{abstract}
Prompt engineering is a technique that involves augmenting a large pre-trained model with task-specific hints, known as prompts, to adapt the model to new tasks. Prompts can be created manually as natural language instructions or generated automatically as either natural language instructions or vector representations. Prompt engineering enables the ability to perform predictions based solely on prompts without updating model parameters, and the easier application of large pre-trained models in real-world tasks. In past years, Prompt engineering has been well-studied in natural language processing. Recently, it has also been intensively studied in vision-language modeling. However, there is currently a lack of a systematic overview of prompt engineering on pre-trained vision-language models. This paper aims to provide a comprehensive survey of cutting-edge research in prompt engineering on three types of vision-language models: multimodal-to-text generation models (\eg Flamingo), image-text matching models (\eg CLIP), and text-to-image generation models (\eg Stable Diffusion). For each type of model, a brief model summary, prompting methods, prompting-based applications, and the corresponding responsibility and integrity issues are summarized and discussed. Furthermore, the commonalities and differences between prompting on vision-language models, language models, and vision models are also discussed. The challenges, future directions, and research opportunities are summarized to foster future research on this topic.
\end{abstract}

}

\maketitle

\IEEEdisplaynontitleabstractindextext

\IEEEpeerreviewmaketitle

\begin{figure*}[!ht]
    \centering
    \begin{subfigure}[b]{0.3\textwidth}
    \centering
        \includegraphics[scale=0.24]{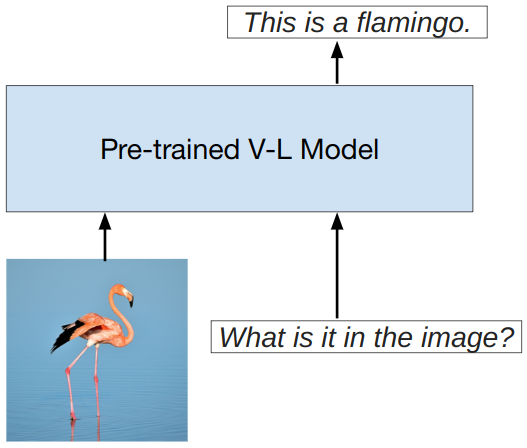}
    \caption{Multimodal-to-Text Generation}
    \label{subf:it2t_gen}
    \end{subfigure} \hspace{3mm}
    \begin{subfigure}[b]{0.3\textwidth}
    \centering
        \includegraphics[scale=0.24]{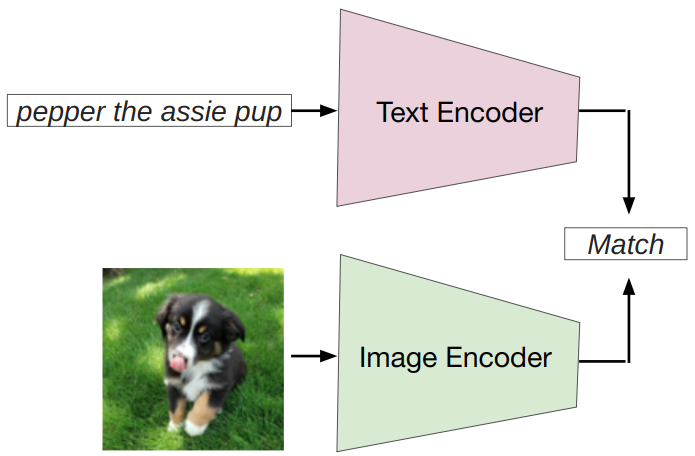}
        \caption{Image-Text Matching}
        \label{subf:i2t_mat}
    \end{subfigure} \hspace{3mm}
    \begin{subfigure}[b]{0.3\textwidth}
    \centering
        \includegraphics[scale=0.24]{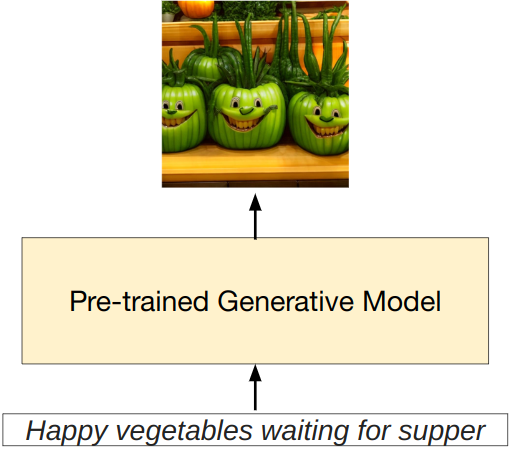}
        \caption{Text-to-Image Generation}
        \label{subf:t2i_gen}
    \end{subfigure}

    \caption{Vision-Language Foundation Models. The cutting-edge research in prompt engineering on Vision-Language Foundation Models is systematically summarized. Three main types of vision-language models are focused in this work, namely, multimodal-to-text generation models (\eg Flamingo~\cite{alayrac2022flamingo}) in subfigure a, image-text matching models (\eg CLIP~\cite{radford2021learning}) in subfigure b, and text-to-image generation models (\eg Stable Diffusion~\cite{rombach2022high}) in the subfigure c. More details of each type are introduced in the later sections.}
    \label{fig:vlm} 
\end{figure*}

\IEEEraisesectionheading{\section{Introduction}\label{Sec:Introduction}}

\IEEEPARstart{P}{rompt} engineering is an approach to adapting a large pre-trained model, also known as a foundation model, to new tasks by augmenting the model input with task-specific hints. Specifically, the model's input is augmented by an additional part, called prompt, which could be manually created natural language instructions~\cite{wei2022chain}, automated generated natural language instructions~\cite{zhang2022automatic}, or automated generated vector representations~\cite{lester2021power}. The natural language instructions have been also referred to as \textit{discrete prompts} or \textit{hard prompts}, while the vector representations are called \textit{continuous prompts} or \textit{soft prompts}. 

Prompt engineering has indeed co-appeared and gained prominence with the emergence of large pre-trained models and together led to a paradigm shift in machine learning (ML). The traditional paradigm requires labeling a considerable amount of data and then training a task-specific ML model from scratch or fine-tuning a pre-trained large model. The model's performance heavily relies on the quality and amount of labeled data, which can be resource-intensive to acquire. Besides, the traditional paradigm requires tuning the model's parameters to some extent, \ie, entire parameters in the case of training an ML model from scratch or fully fine-tuning a pre-trained model and partial parameters in the case of parameter-efficient finetuning. This limits the extensibility of an ML model and requires a specific model copy for each task. Recently, prompting a pre-trained large model to adapt it for specific tasks has become a new trend. The key idea of prompt engineering is to provide hints along with input to guide a pre-trained model for solving a new task using its existing knowledge. If the hints are human-interpretable natural language (\textit{hard prompts}), the related studies have been referred to as \textit{In-Context Learning}~\cite{dong2022survey}, which enable the model to learn from task instructions, demonstrations with a few examples, or supporting information in the context. Also, the hints could be continuous vector representations (\textit{soft prompts}). The related work has been referred to as \textit{Prompt-Tuning}~\cite{lester2021power}, which optimizes prompts directly in the embedding space of the model.

Compared to the traditional paradigm, prompt engineering has multiple advantages. Firstly, it requires a few labeled data to adapt a pre-trained model to new tasks, which greatly reduces the effort of human supervision and computation resource for fine-tuning. Secondly, prompt engineering enables a pre-trained model to perform predictions on new tasks solely based on the prompt without updating any of the model's parameters, allowing serving a large scale of downstream tasks using the same model. This makes it possible to apply large-scale pre-trained models for real-world applications.

Prompt engineering has been first studied and popularized in natural language processing (NLP)~\cite{liu2023pre, qiao2022reasoning}, and then gained great attention in computer vision~\cite{bahng2022exploring, wang2023seggpt}, as well as in vision-language modeling~\cite{alayrac2022flamingo, wu2023visual}. While there is an abundance of literature on prompt engineering in the NLP domain, there is currently no systematic overview available to provide insight into the current state of prompt engineering on pre-trained vision-language models (VLMs), which present their own unique challenges.

In this paper, we aim to bridge this gap by providing a comprehensive survey of cutting-edge research in prompt engineering of pre-trained VLMs. Specifically, we classify prompting methods into two main categories based on the readability of the templates, \ie, hard prompt and soft prompt. hard prompts can be further divided into four subcategories, namely task instruction, in-context learning, retrieval-based prompting, and chain-of-thought prompting. Soft prompts, on the other hand, are continuous vectors that can be fine-tuned using gradient-based methods. 
Note that this survey primarily focuses on prompting methods that maintain the model's architecture, and thus, the methods such as P-tuning~\cite{liu2021gpt} and LoRa~\cite{hu2021lora} that introduce additional modules into the model, are not the primary scope of this survey.

We investigate the prompt engineering on three types of VL models, which are \textit{muiltimodal-to-text generation models}, \textit{image-text-matching models}, and \textit{text-to-image generation models}. A clear definition of each model type is provided in Sec.~\ref{sec:terminology}. 
Moreover, we categorize existing prompt-engineering approaches from an encoder-decoder perspective as shown in Fig.~\ref{fig:vlm}, \ie, encode-side prompting or decode-side prompting, where the prompts are added to the encoder and decoder, respectively.

The rest of this paper is organized as follows. In Sec.~\ref{sec:2-taxonomy}, we summarize and define the taxonomy and notations that we use across this survey. 
Sec.~\ref{sec:3-mm-text},~\ref{sec:4-clip}, and~\ref{sec:5-text-img} present the current progress of prompt engineering on multimodal-to-text generation models, image-text-matching models, and text-to-image generation models, where each section first presents the preliminaries of the corresponding models followed by a detailed discussion of the prompting methods, then investigates the applications and the responsible AI considerations of such prompting methods. Sec.~\ref{sec:unimodal} provides a comparison between prompting unimodal models and VLMs, and we make an in-depth discussion about their analogies and differences. Finally, in Sec.~\ref{sec:7-discussion}, we highlight the challenges and potential research directions. 

In order to facilitate the literature search, we also build and release a project page ~\footnote{https://github.com/JindongGu/Awesome-Prompting-on-Vision-Language-Model/} where the papers relevant to our topic are organized and listed.

\section{Taxonomy}
\label{sec:2-taxonomy}
In this section, terms and notations related to Prompting Engineering on VLMs used throughout the paper are introduced.

\subsection{Terminology}
\label{sec:terminology}
This is a list of terms along with their descriptions. Note that instead of formally defining the following concepts, we provide a general description for readers. 

\vspace{0.1cm}
\begin{itemize}[noitemsep,nolistsep]
\setlength\itemsep{0.5em}
    \item \textit{Prompt:} Additional information or hints provided to a model to guide its behavior or help it perform a specific task;
    \item \textit{Prompting Method:} An approach used to incorporate prompts into the input to guide model behavior or enhance model performance;
    \item \textit{Multimodal-to-Text Generation:} Generating textual descriptions or narratives from multimodal input data, \eg a combination of vision and language data; 
    \item \textit{Image-Text Matching:} Establishing a semantic relationship or alignment between images and textual descriptions;
    \item \textit{Text-to-Image Generation:} Generating visual images from textual descriptions.
    \item \textit{In-context Learning:} A prompting method by providing models with instructions or demonstrations within relevant contexts to solve new tasks without requiring additional training. 
    
    \item \textit{Chain-of-thought:} A prompting method that enhances reasoning skills by instructing a model to generate a sequence of intermediary actions that guide towards solving a multi-step problem and reaching the ultimate solution.

\end{itemize}

\subsection{Notations}
These are the mathematical notations that are followed throughout the paper (Tab.~\ref{tab:notation}). All the formulations of this work will stick to these notations unless otherwise specified. 

\begin{table}[!ht]
    \centering
    \caption{The used mathematical notations are listed. They are followed throughout the paper.}
    \begin{tabular}{c|c}
    \toprule
       $x$  & A clean input image \\
    \midrule
       $t$  & A sentence paired with an image \\
    \midrule
       $y$  & A ground-truth class label of an image \\
    \midrule
       $\chi$ & Input distribution \\
    \midrule
       $f(\cdot)$   & A vision-language model \\
    \midrule
       $f_v(\cdot)$   & A visual encoder \\
    \midrule
       $f_e(\cdot)$   & A textual encoder \\
    \midrule
       $\{v_i\}_{i=1}^M$   &  visual tokens \\
    \midrule
       $\{c_i\}_{i=1}^M$  &  textual tokens \\
    \midrule
        $\{z_i\}_{i=1}^M$   &  visual prompt tokens \\
    \midrule
        $\{t_i\}_{i=1}^M$   &  textual prompt tokens \\
    \midrule
        $H^l$  & $l^{th}$ A layer of the target network \\
    \midrule
        L & Label word token \\
    \midrule
        $H^i_k$ & the $k^{th}$ activation in $l^{th}$ layer of the target model \\
    \midrule
        $z^i$  & Model output logits \\
    \bottomrule
    \end{tabular}
    \label{tab:notation}
\end{table}

\section{Prompting Model in Multimodal-to-Text Generation}
\label{sec:3-mm-text}
\subsection{Preliminaries of Multimodal-to-Text Generation}

Large language models (LLMs) have demonstrated impressive capabilities in the field of NLP, prompting researchers to explore ways of integrating visual modality into these models' training framework. This integration aims to enhance their linguistic prowess and expand their applicability to multimodal tasks.

To maintain consistency with the training methodologies employed by LLMs, generation-based vision-language models (VLMs) typically comprise three essential components: \textit{text feature}, \textit{visual feature}, and \textit{fusion module}. These components synergistically collaborate, enabling the models to effectively leverage textual and visual information to generate coherent and contextually relevant outputs.

Incorporating the visual modality into LLMs has opened up exciting opportunities for various applications, such as visual commonsense reasoning~\cite{huang2023language}, visual question answering~\cite{alayrac2022flamingo, yang2022empirical, tsimpoukelli2021multimodal, huang2023language}, multimodal dialogue systems~\cite{openaigpt4, alayrac2022flamingo, wu2023visual}, \etc. By combining textual and visual cues, VLMs have the potential to provide a more comprehensive understanding of multimodal data and produce outputs that align with human-like reasoning and perception~\cite{vlmsurvey}. Furthermore, the fusion of text and visual features within VLMs plays a crucial role in seamlessly integrating information from both modalities. This fusion process enables the model to capture interdependencies and interactions between textual and visual elements, resulting in more accurate and contextually grounded generations~\cite{vlmsurvey}.

\textbf{Text Feature.} Early studies on VLMs commonly employed the preprocessing technique introduced by BERT~\cite{devlin2018bert}. The raw text undergoes tokenization and is concatenated with special tokens, \texttt{[CLS]} and \texttt{[SEP]}, represented as \texttt{<[CLS],$c_1$,...,$c_m$,[SEP]>}, where token $c_i$ is associated with a word embedding. However, with the progression of language model research, more advanced models have emerged, showcasing emergent abilities such as in-context learning~\cite{pritchett2015learning} and chain-of-thought reasoning~\cite{wei2022chain}. Building upon these advancements, the latest generation of VLMs has embraced powerful language models like T5~\cite{raffel2020exploring} and GPTs~\cite{brown2020language}, which further enhances their linguistic capabilities.

To accommodate different modalities in the input, recent works have introduced new special tokens. For example,~\cite{singh2022flava} incorporate an additional image classification token \texttt{[CLS\_I]}, while~\cite{huang2023language} use \texttt{<image>} and \texttt{</image>} to indicate the beginning and end of the encoded image embedding and \texttt{<s>} and \texttt{</s>} to mark the beginning and end of a sequence. In another approach,~\cite{alayrac2022flamingo} employs \texttt{<BOS>} to represent the ``beginning of sequence" and \texttt{<EOC>} to denote ``end of chunk". These special tokens serve to differentiate and identify the boundaries between different modalities, allowing the model to effectively process and leverage multimodal information.

\noindent
\textbf{Visual Feature.} To obtain a consistent representation of input as a sequence of embeddings for both modalities, the image $x$ is transformed into a sequence of embedding vectors: $x = <v_1, v_2, ..., v_M>$. Accurately representing the information conveyed by images is crucial for downstream tasks but can be challenging. CNN structures have been commonly used in prior research for extracting image features. For instance, models like ViLBERT~\cite{lu2019vilbert} and VL-T5~\cite{cho2021unifying} employ faster R-CNN~\cite{ren2015faster} to detect object regions in images and encode them as a sequence of Region-Of-Interest (ROI) features. However, this approach may overlook important regions in an image. To address this limitation, approaches like OFA~\cite{wang2022ofa} and Flamingo~\cite{alayrac2022flamingo} utilize ResNet to encode information from the entire image, considering a broader context. Additionally, leveraging the powerful feature extraction capabilities of the transformer architecture, models such as SimVLM~\cite{wang2021simvlm}, PaLI~\cite{chen2022pali}, MAGMA~\cite{eichenberg2021magma}, and BLIP2~\cite{li2023blip2} adopt the Vision Transformer (ViT)~\cite{dosovitskiy2020image} architecture for image representation. This allows them to effectively capture visual information and incorporate it into the multimodal framework.

\noindent
\textbf{Fusion Module.} The fusion module plays a crucial role in integrating text and image embeddings to create a joint representation. A well-designed fusion module can capture interactions and relationships between modalities, prevent information loss, avoid semantic mismatch, mitigate biases, and enables comprehensive understanding. For example, in Visual Question Answering (VQA), the fusion module enables the model to leverage both textual and visual information to understand the question and the corresponding image, leading to accurate answers. To improve the ability of answer generation, prompts can be manually designed for different tasks and included as part of the input to the fusion module. These prompts serve as additional information or cues that guide the model's understanding of the question and the image.
As for generation-based VLMs, there are two main types of fusion module approaches based on the integration of visual and textual modalities: \textit{encoder-decoder as a multi-modal fusion module} and \textit{decoder-only as a multi-modal fusion module}.

In the encoder-decoder as a multi-modal fusion module approach, models like VL-T5~\cite{cho2021unifying}, SimVLM~\cite{wang2021simvlm}, OFA~\cite{wang2022ofa}, and PaLI~\cite{chen2022pali} focus on creating a joint representation that combines both modalities at an early stage. The overall formulation can be represented as:

\begin{equation}
\begin{aligned}
y = \mathcal{G}(\mathcal{E}(x_{input}))
\end{aligned}
\end{equation}

\noindent
where the $x_{input}$ represents the given input and $y$ denotes the corresponding ground-truth, respectively. The fusion encoder function $\mathcal{E}$ integrates the visual and textual information to create a joint representation that captures their interactions and dependencies. This fused representation is then fed into the generating module $\mathcal{G}$, which performs further processing and generates the desired outputs for the downstream tasks.

In the decoder-only as a multi-modal fusion module approach, models like Frozen~\cite{tsimpoukelli2021multimodal}, Flamingo~\cite{alayrac2022flamingo}, and MAGMA~\cite{eichenberg2021magma} directly combines the visual and textual information in the decoding stage, without explicitly creating a joint representation at an earlier stage. This approach allows the model to effectively incorporate both modalities during the generation process and produce contextually relevant outputs. The formulation can be represented as:

\begin{equation}
y = \mathcal{G}(x_{input})
\end{equation}

A special case is PICa~\cite{yang2022empirical}, which represents images as textual descriptions and utilizes GPT-3 as the fusion module. This approach treats images as text and leverages a pre-trained language model like GPT-3 to generate outputs based on the pure text input.

In addition, BLIP-2~\cite{li2023blip2} examines the fusion of two distinct modules for integration: the decoder-based OPT~\cite{zhang2022opt} and the encoder-decoder based FlanT5~\cite{chung2022scaling}. The study further offers an analysis of the respective strengths and benefits offered by these fusion modules.

\subsection{Multimodal-Text Prompting Methods}\label{sec:m2t}

\begin{figure*}
    \centering
    \includegraphics[scale=0.36]{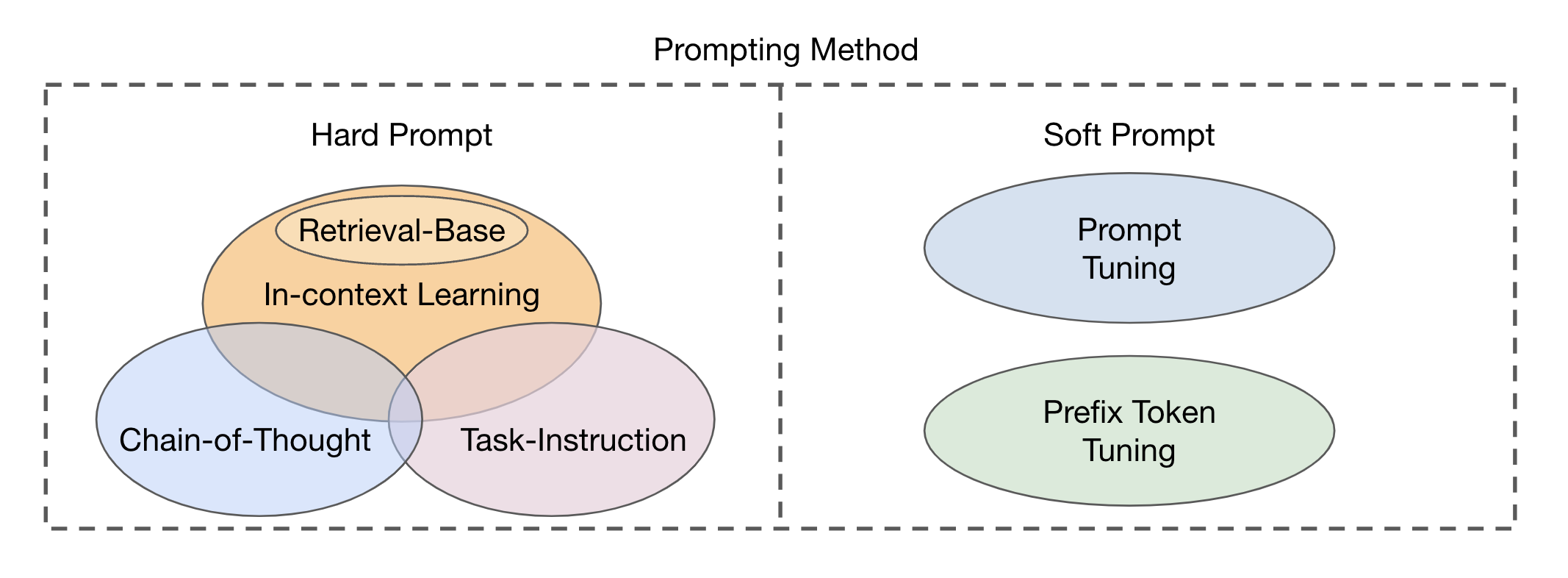}
    \caption{Prompting methods in multimodal-to-text generation. Prompting methods can be divided into two main categories based on the readability of the templates: hard prompt and soft prompt. Hard prompt encompasses four subcategories: task instruction, in-context learning, retrieval-based prompting, and chain-of-thought prompting. Soft prompts are classified into two strategies: prompt tuning and prefix token tuning, based on whether they internally add new tokens to the model's architecture or simply append them to the input. However, this study primarily concentrates on prompt methods that avoid altering the base model, excluding techniques like P-tuning and LoRa that modify the model's core structure.}
    \label{fig:chapt3_prompting_method}
\end{figure*}

Fig.~\ref{fig:chapt3_prompting_method} illustrates the classification of prompting methods. Prompting methods fall into two categories: hard prompts, which are labor-intensive, manually crafted text prompts with discrete tokens, and soft prompts, which are optimizable, learnable tensors concatenated with input embeddings, but lack human readability due to their non-alignment with real word embeddings.

\subsubsection{Hard prompt}
Hard prompts involve manually crafted, interpretable text tokens, \eg adding ``\textit{A photo of }" before the input for captioning tasks. Hard prompts can be further divided into four subcategories: \textit{task instruction}, \textit{in-context learning}, \textit{retrieval-based prompting}, and \textit{chain-of-thought prompting}. It is important to note that retrieval-based prompting is often used to select samples for in-context learning. 

\noindent
\textbf{Task Instruction Prompting.} This method involves the use of carefully designed prompts that provide explicit task-related instructions to guide the model's behavior~\cite{radford2019language,efrat2020turking}. The formulation for this method can be represented as $x_{\text{input}} = \mathcal{H}(x, t)$. Here, $\mathcal{H}$ serves as the task instruction function, taking the image $x$ and text $t$ as inputs and producing the modified input representation $x_{\text{input}}$. 

\noindent
\textbf{In-context Learning.} In-context Learning~\cite{dong2022survey, brown2020language} is a method where the model is exposed to a sequence of related examples or prompts, enabling it to learn and generalize from the provided context. The in-context learning method can be represented using the equation $x_{\text{input}} = \mathcal{H}(\mathcal{C}, x, t)$.  Here, $\mathcal{H}$ denotes the task instruction function which integrates the given context $\mathcal{C}$ with the image $x$ and text $t$ inputs. The resulting modified input representation $x_{\text{input}}$ captures the model's understanding of the context and is used to generate coherent and contextually relevant responses. By exposing the model to a sequence of related examples or prompts, the in-context learning method promotes improved performance in understanding and generating responses~\cite{wei2022chain}.

\noindent
\textbf{Retrieval-based Prompting.} Retrieval-based Prompting~\cite{yang2022empirical, rubin-etal-2022-learning, li-etal-2023-unified, ye2023compositional} is a method that involves selecting prompts or context using retrieval techniques. In this approach, the model retrieves relevant prompts or context from a prompt pool or external knowledge base to guide its generation or decision-making process. The retrieval-based prompting method can be denoted by the formulation: $\mathcal{C} = \mathcal{R}(x,t)$. In this equation, $\mathcal{R}$ signifies the retrieval method that garners pertinent prompts or context based on the image $x$ and text $t$ inputs. The retrieved context $\mathcal{C}$ is then used to guide the model's generation or decision-making process. It is worth noting that the retrieval method $\mathcal{R}$ can vary depending on the specific approach and the available prompt pool or knowledge base. This method allows the model to benefit from existing information and improve its performance by leveraging relevant prompts or context during the generation process~\cite{rubin-etal-2022-learning, li-etal-2023-unified, ye2023compositional}.

\noindent
\textbf{Chain-of-Thought Prompting.} Chain-of-Thought Prompting~\cite{wei2022chain, qiao2022reasoning, zhang2022automatic} is a method where the model is prompted with a series of instructions or questions that progressively build upon each other. Each prompt in the chain adds context or narrows down the focus, enabling the model to generate more coherent and contextually appropriate responses. This method helps the model maintain a logical ``chain" throughout the conversation. The formulation for the chain-of-thought prompting method does not involve a specific equation but rather the iterative process of applying prompts~\cite{wei2022chain}. At each step $l$ in the chain, the model's response from the previous prompt is used as input for the next prompt. This can be represented as $\mathcal{T}^{l+1} = \mathcal{T}^l(x, t)$. Here $\mathcal{T}$ represents the prompt function that takes the image $x$ and text $t$ inputs and generates a response. The output of the $l$-th prompt, denoted as $\mathcal{T}^l(x, t)$, serves as the input for the $(l+1)$-th prompt $\mathcal{T}^{l+1}$. By progressively building upon the previous prompts, the iterative nature of the chain-of-thought prompting method helps the model maintain coherence and generate responses that align with the evolving context of the conversation~\cite{wei2022chain}.

\subsubsection{Soft prompt}
Unlike hard prompts, soft prompts are characterized as continuous vectors that can be fine-tuned using gradient-based methods~\cite{lester2021power, qin-eisner-2021-learning}. For example, this process might involve concatenating a learnable vector with the input embeddings and subsequently optimizing these to align with a particular dataset. Soft prompts can be classified according to whether new tokens are internally incorporated within the model's architecture or simply attached to the input. This distinction generally relates to two specific strategies: \textit{prompt tuning} and \textit{prefix token tuning}. However, this survey focuses exclusively on prompt methods that do not involve modifying the underlying model itself, and thus techniques like P-tuning~\cite{liu2021gpt} and LoRa~\cite{hu2021lora}, which alter the fundamental structure of the model, are not within the primary scope of this study.

\noindent
\textbf{Prompt Tuning.} Prompt tuning~\cite{lester2021power} creates continuous vector representations as input hints. During the training process, the model learns to refine the prompts, aiming to improve its performance on specific tasks. This method enables the model to dynamically generate effective prompts based on its understanding of the task. The objective of prompt tuning, with the prompting parameter $x_p$, can be demonstrated as follows:
\begin{equation}  
    \operatorname*{argmin}_{x_p} \mathcal{L}(\mathcal{F}(y_i,x_p)|y_{\textless i}, x_{input})
\end{equation}
where $\mathcal{F}(y_i,x_p)$ represents the model's output given the prompting parameter $x_p$. Here $y_{\textless i}$ denotes the previously generated outputs, and $x_{\text{input}}$ refers to the modified input based on the prompt. The objective of prompt tuning is to minimize the loss $\mathcal{L}$ between the model's output and the desired output, given the previously generated outputs and the modified input. By continuously refining the prompts through prompt tuning, the model adapts its behavior and improves its performance on specific tasks. The dynamic generation of effective prompts based on the model's understanding enhances its capability to generate accurate and contextually relevant responses. 

\noindent
\textbf{Prefix Token Tuning.} Similar to prompt tuning, prefix token tuning~\cite{li2021prefix} involves adding task-specific vectors to the input. However, in this case, these vectors are inserted in all model layers and can be trained and updated independently while keeping the rest of the pre-trained model's parameters frozen.

It's worth noting that these prompting methods are not mutually exclusive. They can be combined and used together to achieve desired results in various settings and tasks. The choice of prompting method depends on the specific task, dataset availability, and the desired level of control and customization required for the model's behavior.

\subsection{Advances in Prompting Techniques for VLM}
This section will overview the use of prompting techniques in various VLMs to boost performance. For a clear and structured presentation, models will be divided into two main types based on their fusion modules: 1) models utilizing an encoder-decoder as the fusion module, and 2) models employing a decoder-only as the fusion module.

\noindent
\textbf{Prompting Models with Encoder-decoder as the Fusion Module.} Early studies in VLMs often involved designing task-specific architectures on top of transformer encoders. However, recent advancements have introduced a unified vision-language framework that incorporates an encoder as the fusion module. Notable examples of such models include VL-T5~\cite{cho2021unifying}, SimVLM~\cite{wang2021simvlm}, and OFA~\cite{wang2022ofa}. They employ two main prompting methods: hand-crafted instructions and prompt tuning.

Both VL-T5~\cite{cho2021unifying} and OFA~\cite{wang2022ofa} utilize text prefixes as prompts. For example, ``\textit{vqa:}" is used for vision question answering, and ``\textit{caption:}" is employed for image captioning tasks. SimVLM~\cite{wang2021simvlm} introduces the prefix ``\textit{a photo of:}" to enhance the quality of decoded captions. In addition, VL-T5~\cite{cho2021unifying} introduces shared visual sentinel tokens (\texttt{<$vis\_i$>}) to specify corresponding image regions of Region of Interest (RoI) features. Text sentinel tokens (\texttt{<$text\_i$>}) are used to replace contiguous text. Similarly, OFA~\cite{wang2022ofa} generates location tokens that specify the position of the region (\texttt{<$x_1$,$y_1$,$x_2$,$y_2$>}). These special tokens facilitate the structured incorporation of visual and textual information.

Building upon these special tokens, VL-T5~\cite{cho2021unifying} utilizes the prompt ``\textit{caption region: \texttt{<$vis\_i$>}}" for the grounded captioning task, indicating that the model should generate a caption based on the specified visual region. OFA~\cite{wang2022ofa} prompts the proposed grounded question answering task using the template ``\textit{Q: what color is the car in the region? region: \texttt{<$x_1$,$y_1$,$x_2$,$y_2$>} A:}", providing instructions for the model to answer the question by referring to the specified visual region. Prompt tuning on OFA is explored by~\cite{yang2022prompt}, who introduce tunable prompt embeddings at each layer. Experimental results demonstrate that this lightweight prompt-tuning approach is not only efficient but also resilient against adversarial attacks.

\noindent
\textbf{Prompting Models with Decoder-based Fusion Module.} Another line of research focuses on utilizing the decoder as a fusion module in VLMs. Frozen~\cite{noever2023multimodal} and BLIP-2~\cite{li2023blip} exemplify models that employ image conditional prefix tuning. Frozen~\cite{tsimpoukelli2021multimodal} introduces the concept of preserving the language capabilities of a LLM while incorporating visual information as a prefix. It achieves this by freezing the model and training a separate vision encoder to represent images. In Frozen, visual information is represented as a sequence of two embeddings, serving as a visual prefix. The authors also propose task induction techniques, such as instructing the model to ``\textit{Answer with dax or blicket,}" and evaluate the model's performance with various forms and amounts of in-context learning for downstream tasks. To effectively facilitate cross-modal alignment, BLIP-2~\cite{li2023blip} does not fine-tune the vision encoder. Instead, it introduces a Querying Transformer (Q-Former) to extract visual features from the frozen image encoder, using the extracted query embeddings as soft visual prompts. MAGMA~\cite{eichenberg2021magma} follows a similar approach to Frozen, incorporating a new image prefix encoder while keeping the language model frozen. Task instructions, such as ``\textit{A picture of }" are used for image captioning. Flamingo~\cite{alayrac2022flamingo} explores the capabilities of few-shot learning and employs various prompt techniques. The authors introduce special tokens, \texttt{<BOS>} (beginning of sequence) and \texttt{<EOC>} (end of chunk), to differentiate sample pairs. In the zero-shot scenario, text prompts that do not contain corresponding vision information are used. In the few-shot setting, different formatting is employed for various tasks (\eg ``\textit{Question: \{question\} Answer: \{answer\}}" for visual question-answering tasks), and the retrieval-based in-context example selection (RICES)~\cite{yang2022empirical} approach is utilized to select suitable sample pairs as prompts. Prompt ensembling techniques are also employed to calculate the final scores. For specific tasks such as HatefulMemes, prompts are designed to incorporate provided OCR information. Additionally, hand-crafted dialogue prompts are specifically designed for presented dialogues.
\cite{chen2022pali} extends the multilingual capabilities of LLMs to VLMs without freezing any parameters. They achieved this by explicitly specifying the intended language in the prompt instruction. For example, a prompt may be formulated as ``\textit{Generate the alt\_text in \texttt{<lang>}}", where \texttt{<lang>} represents the language code associated with the desired text string. Furthermore, Microsoft proposes a series of Multimodal Large Language Models (MLLM), namely Kosmos-1~\cite{huang2023language} and Kosmos-2~\cite{peng2023kosmos}. These models possess the ability to perceive diverse modalities and evaluate a wide range of tasks, including zero-shot, few-shot, and multimodal chain-of-thought prompting scenarios. Textual instructions are used to enable the model to better understand downstream tasks. For example, in Kosmos-1~\cite{huang2023language} phrases like ``\textit{Here are three/four/eight images:}" and ``\textit{The following image is:}" are employed for the Raven IQ test. In chain-of-thought prompting, Kosmos-1 first uses a prompt (\eg ``\textit{Introduce this picture in detail:}") to guide the model to generate a rationale. Then, a task-aware prompt incorporating the generated rationale is utilized to produce the final results. Based on Kosmos-1~\cite{huang2023language}, Kosmos-2~\cite{peng2023kosmos} incorporating grounding and referring capabilities by using text span with bounding box as prompt, \ie, ``\textit{\texttt{<p>} text span \texttt{</p><box><loc1><loc2></box>}}, where \texttt{<loc1>} and \texttt{<loc2>} are location tokens  \texttt{<p>}, \texttt{</p>}, \texttt{<box>} and \texttt{</box>} are special boundary and text span tokens, respectively.

PICa~\cite{yang2022empirical} takes a different approach by not learning visual embeddings. Instead, it converts images into textual descriptions and queries GPT-3 directly to predict the answer. Leveraging the few-shot learning ability of GPT-3, PICa adapts to the visual question-answering (VQA) task with only a few in-context examples during inference time. GPT-4~\cite{openaigpt4}, the latest version of ChatGPT, has been introduced as an advanced VLM. In addition to employing task-specific hard prompts, GPT-4 also incorporates the in-context learning approach to tackle complex tasks such as AP Art History~\cite{Nici_2020}.

\subsection{Understanding Prompting}
To deeply understand the factors impacting prompting in multimodal-to-text generation models, the following aspects will be introduced:

\noindent
\textbf{Dataset-specific Prefixes.} The choice of text prompts can have a significant impact on the performance of models. VL-T5~\cite{cho2021unifying} experimented with a single prefix ``\textit{vqa}" for both Visual Question Answering (VQA) and GQA~\cite{hudson2019gqa} tasks. The results demonstrated that a single model can effectively handle multiple VQA tasks without the need for dataset-specific prefixes.

\noindent
\textbf{Freezing the Language Model.} Many explorations of prompting in multimodal-to-text generation models rely on the powerful generative capabilities of language models. To preserve the extensive capabilities of LLMs, approaches like Frozen~\cite{tsimpoukelli2021multimodal}, MAGMA~\cite{eichenberg2021magma}, Flamingo~\cite{alayrac2022flamingo}, and BLiP2~\cite{li2023blip2} freeze the language model during training. This prevents knowledge loss and enables the retention of prompt capabilities. On the other hand, approaches like OFA~\cite{zang2022unified} and KOSMOS-1~\cite{huang2023language} directly adopt the encoder-decoder structure without additional model components to pursue unified models. However, fine-tuning the language model alone can lead to a decrease in language ability. To address this, both OFA~\cite{zang2022unified} and KOSMOS-1~\cite{huang2023language} add language-only tasks during training to prevent the loss of language ability.

\noindent
\textbf{In-context Learning.} Recent studies have demonstrated that the in-context learning capabilities of language models can be successfully transferred to vision-language-generating models. Frozen~\cite{tsimpoukelli2021multimodal} exhibits the capability of fast concept binding, enabling the model to associate a new word with a visual category using only a few examples and immediately utilize that word appropriately. While the model performs well in the two-way setting (two new words), this ability fails to transfer to the five-way setting (five new words). Experimental results also indicate that increasing the number of in-context learning samples enhances model performance, but there is a saturation point, and additional repeated content can even lead to a decline in performance. Similar conclusions have been drawn in the case of Flamingo~\cite{alayrac2022flamingo}. Both Flamingo~\cite{alayrac2022flamingo} and Kosmos-1~\cite{huang2023language} demonstrate that employing individual text prompts instead of image-text pairs can improve model performance. However, it is important to note that using individual text prompts can introduce bias to the model~\cite{alayrac2022flamingo}.

\noindent
\textbf{Prompt Tuning.}~\cite{yang2022prompt} conducted a study on Prompt tuning in generative multimodal models. Their findings indicate that prompt tuning consistently exhibits greater robustness than finetuning across various tasks. The study also highlights the impact of different setups on prompting performance, revealing that longer prompts with more parameters can facilitate improvements. However, there is a diminishing marginal utility, and excessively long prompts may even have a detrimental effect on performance. Furthermore, the results suggest that inserting prompts at the bottom layers might lead to better performance.

\subsection{Application of Prompting}
Prompting has been widely adopted in many vision-language tasks evolving text generation, demonstrated promising results, and inspired a new learning paradigm, \ie, in-context learning. 

\noindent
\textbf{Visual Question Answering.} The goal of visual question answering (VQA) is to train models to understand the information in an image and answer questions about it in natural language. In-context prompts show surprising results in few-shot~\cite{alayrac2022flamingo, yang2022empirical, tsimpoukelli2021multimodal, huang2023language} and zero-shot scenarios~\cite{wang2021simvlm, chen2022pali, alayrac2022flamingo, huang2023language}.  Some work also applies prompts to web page question answering~\cite{huang2023language} and grounded question answering~\cite{wang2022ofa}. Web page question answering aims to find answers to questions from web pages which requires comprehension of both the semantics and structures of texts. Huang \etal~\cite{huang2023language} uses the template prompt \textit{``Given the context below from the web page, extract the answer from the given text like this: Question: Who is the publisher of this book? Answer: Penguin Books Ltd. Context: \{WebText\} Q: \{question\} A: \{answer\}"} where \textit{\{WebText\}} stands for the text extracted from web pages. Grounded question answering is firstly designed to reflect the strong transferability of the One For All (OFA) model~\cite{wang2022ofa}. In this task, the model should answer a question about a certain region, and special region tokens for hard prompts are designed. 

\noindent
\textbf{Visual Commonsense Reasoning.} This task requires an understanding of the properties of everyday objects in the real world, such as object size reasoning and object color reasoning\cite{huang2023language}. The model is required to predict the size or color relation between  The Kosmos model~\cite{huang2023language} uses example prompts like \textit{Is \{Item1\} larger than \{Item2\}? \{Answer\}} and \textit{The color of \{Object\} is? \{Answer\}} in the zero-shot scenarios and achieves promising results.

\noindent
\textbf{Zero-shot Image Classification.}
Prompting combined with large-pre-trained multimodal models has shown great transferability on out-of-domain test data such as zero-shot image classification. Kosmos~\cite{huang2023language} concatenates the input image with a prompt like \textit{The photo of the} and lets the model complete the prompt sentence with the predicted class. Besides, to incorporate additional rules in the classification, Kosmos also sends class descriptions along with prompts to prompt the model for a specific category. 

\noindent
\textbf{Image Captioning.}
Generating descriptions given an image is a typical multimodal-to-text generation task that requires the comprehension of both vision and language information. Prompts are used mostly in few-shot and zero-shot scenarios and demonstrate powerful capacity. Flamingo~\cite{alayrac2022flamingo} and PaLI~\cite{chen2022pali} adopt prompts to generate image captions in few-shot settings. For example, PaLI~\cite{chen2022pali} uses the prompt \textit{Generate the alt\_text in EN} to generate image captions. Prompts are also studied in zero-shot settings, such as BLIP-2\cite{li2023blip2}, MAGMA~\cite{eichenberg2021magma}, SimVLM~\cite{wang2021simvlm}, and OFA~\cite{wang2022ofa}.

\noindent
\textbf{Chatbot.} The advent of chatbots such as ChatGPT~\cite{chatgpt} is one of the most remarkable breakthroughs in AI research. Following work such as Visual ChatGPT~\cite{wu2023visual} and GPT4~\cite{openaigpt4} extend chatbots to multimodal applications which support both images and text prompts. Visual ChatGPT~\cite{wu2023visual} is built based on ChatGPT and visual foundation models. It uses a Prompt Manager which specifies input-output formats, converts visual information to language format, and handles histories of different visual foundation models. GPT4~\cite{openaigpt4} is able to accept prompts consisting of both images and texts, which lets users specify any vision and language task by generating text outputs given arbitrarily interlaced text and image prompts. Besides, some work migrates GPT to a specific domain such as BiomedGPT on biomedical research~\cite{zhang2023biomedgpt}.

\subsection{Responsible AI Considerations of Prompting}
\label{sec:m2t-ethical}
Language-based VLMs inherit the risks of the underlying LLMs and vision models, such as gender and racial biases when prompted with images~\cite{weidinger2021ethical}. Several surveys on the ethics of LLMs are available~\cite{weidinger2021ethical, guo2022threats}. Some work studies the robustness of VLMs against both natural distribution shifts~\cite{qiu2022benchmarking} and adversarial robustness~\cite{zhao2023evaluating}. A recent study~\cite{chen2023benchmarking} investigates the robustness of prompt tuning on VLMs against natural distribution shifts. Moreover,~\cite{gu2023towards} proposes robust prompt tuning on VLMs by integrating multiple-scale image features into the prompt.

\section{Prompting Model in Image-Text Matching}
\label{sec:4-clip}

\begin{figure*}
    \centering
    \includegraphics[width=\textwidth]{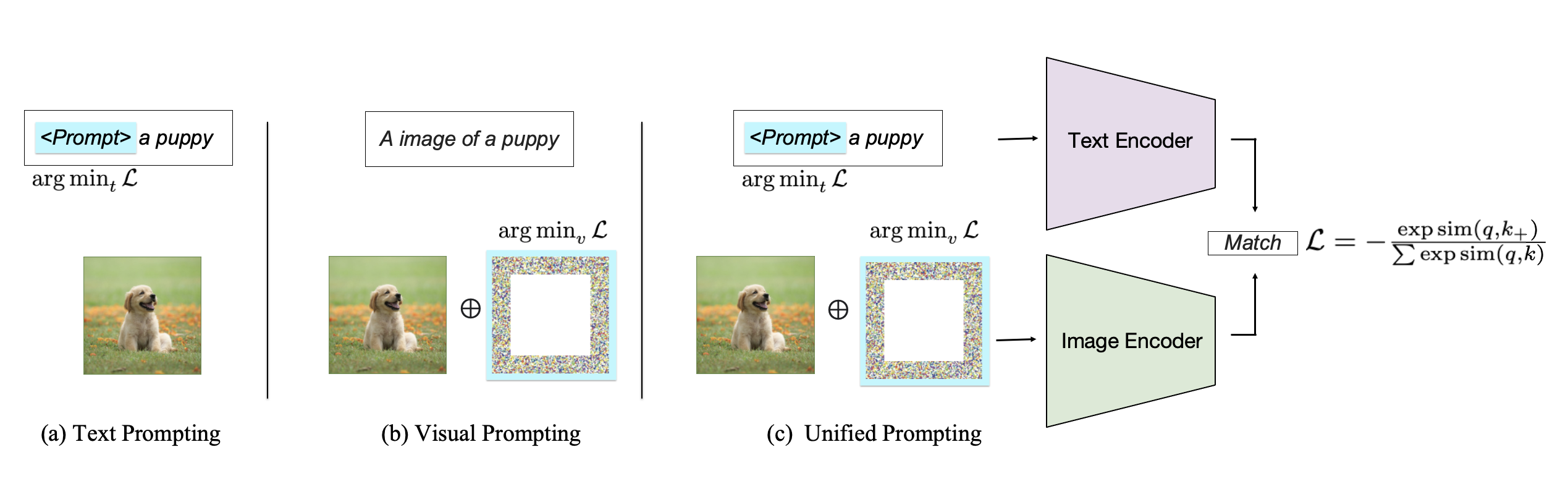} 
    \caption{Prompting tuning on Image-Text Matching VLMs can be applied in different branches: (a) Text prompting, (b) Visual prompting, and (c) Unified prompting, on the input data. Light blue boxes denote learnable parameters.  A matching loss is employed to optimize over a small number of learnable parameters, in the loss formulation $q$ denotes the querying modality and $k$ denotes the target modality.} \vspace{-0.2cm}
    \label{fig:branch}
\end{figure*}

\subsection{Preliminary of Image-Text Matching Models}

Matching-based VLMs have introduced a novel training paradigm that facilitates the acquisition of joint multi-modal representations. Prominent models in this field, such as CLIP~\cite{radford2021learning}, ALIGN~\cite{li2021alignb}, ALBEF~\cite{li2021alignb} and Multi-Event CLIP~\cite{geng2022meclip}, leverage contrastive learning techniques to achieve joint representations for images and texts with a learning objective that aims to bring the representation of an image-text pair closer together while pushing non-pairs further apart.

By expanding training datasets~\cite{radford2021learning} and scaling up the model parameter, matching-based models exhibit adaptability across a broad spectrum of downstream tasks, including zero-shot benchmarks and fine-tuning scenarios.

Depending on the target of prompting, existing methods can be classified into three categories: \textit{prompting the text encoder}, \textit{prompting the visual encoder}, or \textit{jointly prompting both branches} as shown in Fig.~\ref{fig:branch}. These approaches aim to enhance the flexibility and task-specific performance of VLMs in recent studies.

A classic matching loss is formulated as below to align the image and text embeddings with an Image-to-Text loss $\mathcal{L}_{i2t}$ and a Text-to-Image loss $\mathcal{L}_{t2i}$.

% some equations
\begin{equation}
    \mathcal{L}_{i2t} = - \frac{1}{N}\sum^{N}_{i}\log(\frac{\exp{\mathrm{sim}(f_v^l(v_i), f_t^l(t_i))}}{\sum^{N}_{j}\exp{\mathrm{sim}(f_v^{l}(v_i), f^{l}_t(t_j)}})
\end{equation}

\begin{equation}
      \mathcal{L}_{t2i} = - \frac{1}{N}\sum^{N}_{i}\log(\frac{\exp{\mathrm{sim}(f_t^l(t_i), f_v^l(v_i))}}{\sum^{N}_{j}\exp{\mathrm{sim}(f_t^{l}(t_i), f^{l}_v(v_j)}})
\end{equation}

By prompting, we substitute model input by following learnable prompts:
for textual prompts:
$
    f^{l}_{v}([\{v_i\}^{M}_{i=1},\{z_i\}^{M}_{i=1}])
$
and for the general visual prompt:
$    f^{l}_{t}([\{c_i\}^{M}_{i=1},\{t_i\}^{M}_{i=1}])$
where $M$ is the number of prompts we use.

\subsection{Prompting Text Encoder of VLM}
Prompting language models has been long studied. 
As discussed in Sec.~\ref{sec:m2t}, we categorize prompts into hard prompts and soft prompts in this section.
As shown in Fig.~\ref{fig:branch}(a), learnable textual prompts are optimized on image-text pairs in a supervised manner. Recent works~\cite{huang2022unsupervised} also investigate a different scenario with unlabelled data.
In this section, we will delve into the details of soft prompts, exploring different types, such as global prompts, task-specific prompts, and instance-specific prompts.

\subsubsection{Hard prompt}
Prompting language models within the context of VLMs has been extensively investigated. The introduction of prompts has played a pivotal role in discovering and utilizing large-scale pre-trained Language Models.
Textual prompts mitigate handcrafted text templates (\eg,``\textit{a photo of a [CAT]}"), 
which enables the model to understand and respond to specific tasks without requiring explicit task-specific training, showcasing the flexibility and versatility of the model.
\cite{radford2021learning} utilize hard prompts to test its zero-shot performance on several tasks.
Hard prompts demand significant expertise in the domain and often involve high costs. This has given rise to a new learning paradigm as carefully refining prompts to optimize performance.

\subsubsection{Soft prompt}
The task of selecting an appropriate prompt is a complex endeavor that demands experience and domain expertise, and significantly impacts model performance. This raises an important question: can we dynamically 'search' for optimal prompts using gradient-descent-based learning methods?
Soft prompts refer to prompts that incorporate learnable parameters within their design. We categorize soft prompts into three main types: global soft prompts, group-specific prompts, and instance-specific prompts.

\noindent
\textbf{Global Soft Prompt.}
A straightforward yet powerful approach to adapting language models for downstream tasks involves modifying template tokens specifically for those tasks. Studies such as~\cite{gao2021making, shu2022test,zhou2022learning} have employed learnable prompts as input token embeddings when dealing with new tasks. Compared to fine-tuning the entire model, learning a small set of prompt embedding parameters proves to be more parameter-efficient and data-efficient.
These prompts, referred to as ``global soft prompts," are utilized consistently across all instances within a given task. The term ``global" signifies their universal usage throughout the task, enabling the model to generalize and perform well across inputs. 

\noindent
\textbf{Group-specific Prompt.}
Several recent studies~\cite{zhou2022learning, ju2022prompting, shen2022multitask} have employed a group of soft prompts specifically tailored to adapt to different tasks or types of inputs. These models enable the models to query and select appropriate prompts dynamically.
\cite{zhou2022learning,ju2022prompting,shen2022multitask} use a group of soft prompts targeted to adapt different tasks/types. Different prompts are queried based on the input data.
CoOp~\cite{zhou2022learning} finds that using different context prompts for classes (class-specific context) can enhance performance in fine-grained classification.  
\cite{ju2022prompting} uses task-specific prompts to adapt CLIP on a wide range of video understanding tasks.
\cite{shen2022multitask} proposes to use MVLPT with different task prompts for source and target tasks to share knowledge across task-specific prompts.

\noindent
\textbf{Instance-specific Prompt.}
While effective in some cases, task-grouped prompts can suffer from overfitting issues and may struggle to adapt to unseen classes or novel samples. In contrast, instance-specific prompts aim to customize prompts for individual samples, allowing for a more personalized and adaptive approach.
CoCoOp~\cite{zhou2022conditional} is a model that adopts instance-adaptive prompts, specifically instance-specific prompts, instead of relying solely on global prompts. This approach has been shown to enhance the generability of the model.

\subsection{Prompting Image Encoder of VLM}
In line with the achievements of prompt tuning in Natural Language Processing, there have been endeavors to extend the concept to visual inputs. According to the way of designing visual prompts, we categorize them into two classes: patch-wise prompts, where prompts are added as visual patches that are prepended to the original images, and annotation prompts which involve annotating prompts directly on the raw images

\noindent
\textbf{Patch-wise Prompts.}
Adding learnable patches as visual prompts is an intuitive method to incorporate visual cues into VLMs. Just as textual soft prompts serve as input tokens,~\cite{jia2022visual} introduces Visual Prompt Tuning (VPT), which learns a small set of visual prompts as visual patches. These patches are concatenated with input images to adapt pre-trained models to new tasks.
VPT investigates visual prompts in the input and latent layers and outperforms most other adaptation methods like full fine-tuning. 
In a similar vein,~\cite{bahng2022exploring} explores the use of visual perturbation as a visual prompting technique. Through adversarial reprogramming, the model learns to add visual prompts to input images.
Additionally,~\cite{shen2022multitask} adopts patchified visual tokens as learnable input embeddings.
\cite{wu2022unleashing} proposes applying normalized visual prompts to augmented images, unleashing the potential of visual prompting in diverse data settings.
In terms of promoting diversity in prompts,~\cite{huang2023diversity} employs different visual prompts for distinct subsets of data.

\noindent
\textbf{Annotation Prompts.}
Visual prompting can also be performed explicitly by directly manipulating images, similar to the process of annotation, which we term annotation prompts.
Colorful Prompt Tuning (CPT) introduced in~\cite{yao2022cpt} focuses on colorizing specific regions of images as visual prompts. By incorporating color cues, the model is guided to ground objects and better understand the visual context.
\cite{shtedritski2023what} explores the use of annotations, such as red circles, as an innovative visual prompting design. These annotations serve as cues to guide the model's attention toward specific areas of interest, thereby enhancing its understanding of images. The study delves into CLIP's emergent ability to comprehend images through the clever use of visual prompts.
Furthermore,~\cite{bar2022visual} proposes the use of example input and output images as visual prompts. By providing a pair of images demonstrating a desired task, such as image inpainting, edge detection, or image colorization, the model is guided to complete similar tasks based on the provided examples.
These studies demonstrate the creative and effective use of explicit visual prompting methods and offer a practical and interpretable approach to improving the model's performance.

\subsection{Unified Prompting on VLM}
As prompt engineering continues to advance in both the vision and language branches, there has been a recent development in joint prompting. This approach aims to enhance matching-based VLMs by leveraging prompts from both the visual and language domains. As in Fig.~\ref{fig:branch}, learnable prompts in both branches are optimized.
According to whether the visual prompt and textual prompt are independent of each other, they can be categorized into coupled and decoupled unified prompting, respectively, in out follow-up discussion.

\noindent
\textbf{Coupled Unified Prompting.}
UPT~\cite{zang2022unified} issues that prompting single modality does not fit all cases:  textual prompts may struggle to handle data with high intra-class visual variance, while visual prompts may struggle with data exhibiting high inter-class visual variance. Thus it employs a tiny neural network to optimize learnable textual and visual prompts jointly and finds unified prompting outperforms any unimodal prompting.

\noindent
\textbf{Decoupled Unified Prompting.}
\cite{shen2022multitask} employs soft textual prompts and VPT-like visual prompts on both the language and vision branches. This approach leverages the benefits of prompt engineering in both modalities
~\cite{yao2022cpt} introduces an innovative approach that combines visual and textual sub-prompts for visual grounding tasks. By utilizing regions in the image as visual prompts and phrases in a sentence as textual sub-prompts, the model can establish co-reference across different modalities. 
\cite{khattak2023maple} introduces MaPLe hierarchical prompts on both branches and synergizes prompt training in both modalities via a Vision-Language coupling function.
By leveraging the strengths of both visual and textual prompts, joint prompting contributes to more effective and versatile VLMs that excel in multimodal understanding.

\subsection{Application of Prompting}
Prompting matching-based VLMs offers the promise of transferring representations learned by pre-trained models to downstream domains and niche tasks like pure-vision tasks including image/video classification, semantic segmentation, relation detection, and multimodal tasks.

\noindent
\textbf{Image Classification.}
Image classification is extensively researched in computer vision for many years. A new approach to object classification has been suggested by prompting the text encoders in VLMs.
A Na\"ive solution to image classification is using a fixed prompt like ``\texttt{An image of [CLASS]}" as in CLIP~\cite{radford2021learning} and TPT~\cite{shu2022test},  which uncovers pre-trained capacities of zero-shot classification performance. 
Accordingly, learnable prompts are adapted to image classification in works like~\cite{shu2022test}
Prompt engineering also shows its efficacy in more challenging classification tasks like long-tailed classifcation~\cite{dong2023lpt}, multi-label classification~\cite{guo2023texts,sun2022dualcoop}.

\noindent
\textbf{Text Classification.}
Text classification appears to present a dual challenge akin to that of image classification. Due to the scope of this survey, we only cover works that focus on text classification of VLMs.
\cite{wen2022visual} uses visual prompts concerning different classes to better leverage visual information for text classification.

\noindent
\textbf{Object Detection.}
Object detection is aimed at predicting class labels of object bounding boxes in an image.
With abundant information on classes in texts, prompt engineering is also used for multi-label recognition.
\cite{sun2022dualcoop} proposes Dual Context Optimization (DualCoOp) using labels as a part of prompts and learning positive and negative prompt pairs to align images and prompts to solve multi-label recognition tasks.
\cite{guo2023texts} proposed Texts-as-Images (TaI) prompting for multi-label detection.
Open-vocabulary object classification is a promising application of prompt engineering in object classification, where the detectors can predict new classes that are not in training. ViLD~\cite{gu2022openvocabulary} generate a fixed prompt template, \eg ``\textit{a photo of [CATEGORY] in the scene}".~\cite{du2022learning} introduces detection prompt (DetPro) to learn continuous prompt representations. 
PromptDet\cite{feng2022promptdet} uses regional prompt learning to align region features and text features.

\noindent
\textbf{Visual Relation Detection.}
Visual relation detection is a computer vision task that targets extracting relations between objects in an image.
Prompt tuning boosts visual relation detection with its powerful commonsense knowledge contained in LLMs.
\cite{xiao2022optimizing} optimizes a small continuous task-specific vector for visual relation detection.
\cite{he2022openvocabularya} pre-trains VLMs with a matching-based strategy to align image regions and dense captions and fine-tune a decoder with soft prompts to generate relation predictions.
\cite{gao2023compositionala} presents a Relation Prompt for video open-vocabulary relation detection by generating subject-object sensitive prompts based on object motion cues. 

\noindent
\textbf{Semantic Segmentation.}
Semantic segmentation is a classic computer vision task with the goal of assigning each pixel to a class label.
DenseCLIP~\cite{rao2022denseclip} converts image-text matching to pixel-text matching to enable a pixel-wise dense prediction including semantic segmentation task; class-conditioned text prompts are used to contextualize visual cues in texts.
Segment anything~\cite{kirillov2023segment} presents a large-scale foundation model for segmentation which takes images and promptable segmentation queries as inputs.

Not only on a single task but prompting has also been proven to be beneficial for domain adaptation and generalization of pre-trained models. Further studies investigate prompt tuning on pre-trained model transferability under distribution shift.

\noindent
\textbf{Domain Adaptation.}
Prompt learning also enables continual learning of pre-trained models in tasks like test-time domain adaptation, which aims to adapt models to unlabeled test data under a distribution shift.
\cite{ge2022domain} attempts to embed domain information discrepancy in domain-specif textual prompts and can preserve semantic features of pre-trained VLMs. 
\cite{gao2022visual} adds prompts to different stages of ViT and fine-tunes prompts in the unlabelled target domain.

\noindent
\textbf{Continual Learning.}
Continual learning is aimed at tackling catastrophic forgetting in non-stationary data distribution. Prompt tuning becomes a new methodology for continual learning.
Learning to Prompt (L2P)~\cite{wang2022learning} shows a brand new prompt-based approach for continual learning by querying trainable task-specific prompts from a prompt pool for each input instance and prepends it to input before pre-trained models to instruct the model.
\cite{wang2022dualprompt} presents DualPrompt, to learn task-invariant and task-specific instructions across tasks, unlike that L2P uses only one prompt tool and calls them General and Expert prompt space.

\noindent
\textbf{Domain Generalization}
Domain generalization targets adapting models to unseen domains in the training stage. 
\cite{zheng2022prompt} encapsulates domain-specific knowledge in domain prompts generated by a prompt adapter and prepends it with input data; at test-time, prompts are generated based on the similarities between domains.

\subsection{Responsible AI Considerations of Prompting}
The importance of AI Integrity and ethics has been attached to prompting matching-based VLMs to construct trustworthy multimodal models. The discussion covers model robustness, safety, fairness, bias, privacy, and the like.

\noindent
\textbf{Adversarial Robustness of Prompt.}
Robustness analysis evaluates the performance of the model under different conditions and perturbations.
\cite{mao2023understanding} studies how VPT and fine-tuning improve zero-shot robustness under adversarial attack on CLIP and finds that VPT is more effective in the absence of texts. 
\cite{chen2023visual} also attempt to leverage universal visual prompting to improve the adversarial robustness at test time. Visual prompting is more flexible compared to conventional adversarial defenses, as it allows universal (\ie, data-agnostic) input prompting templates, which are capable of plug-and-play during testing. 
\cite{fang2022data, shi2023effective} investigate the reasons behind VLMs' robustness to natural distribution shifts systematically and reveals that diverse training data is the primary reason for robustness gain.
\cite{xu2022exploring} explores the model vulnerability that injecting triggers brings to pre-trained models in prompt tuning.

\noindent
\textbf{Backdoor Attack of Prompt Learning.}
\cite{carlini2022poisoning} studies on the backdoor and poisoning attacks on CLIP and find CLIP trained on manually labeled data suffer badly from such attacks. It shows that the training on noise and uncurated datasets makes backdoor and poisoning attacks a significant threat. 
\cite{jia2022badencoder} proposes a new backdoor attack method named BadEncoder on CLIP and exposes this threat to VLMs. Once a pre-trained image encoder has been injected backdoors, the downstream classifiers built on it for different downstream tasks simultaneously inherit the backdoor behavior. 
Given such vulnerability to backdoor attacks, CleanCLIP~\cite{bansal2023cleanclip} is proposed as a finetuning framework that weakens the learned spurious associations introduced by backdoor attacks. 

\noindent
\textbf{Fairness and Bias.}
Social bias is an important topic in a fair AI system. A wide range of works have studied different aspects of biases.
\cite{agarwal2021evaluating} showcases an analysis of bias regarding race and gender misclassification in the CLIP model. 
In the meantime, many existing works focus on de-biasing the model.
In particular,~\cite{chuang2023debiasing} attempts to alleviate bias by calibrating the biased prompted texts to debiased content while~\cite{kong2023mitigating} proposes to mitigate biased results in image retrieval tasks by post-processing of the VLMs output.
In addition,~\cite{smith2023balancing} introduces a new dataset debiasing pipeline to augment the dataset with healthy data.

\section{Prompting Model in Text-Image Generation}
\label{sec:5-text-img}
\subsection{Preliminary of Text-Image Generation Models}
\begin{figure}[t]
    \centering
    \includegraphics[scale=0.4]{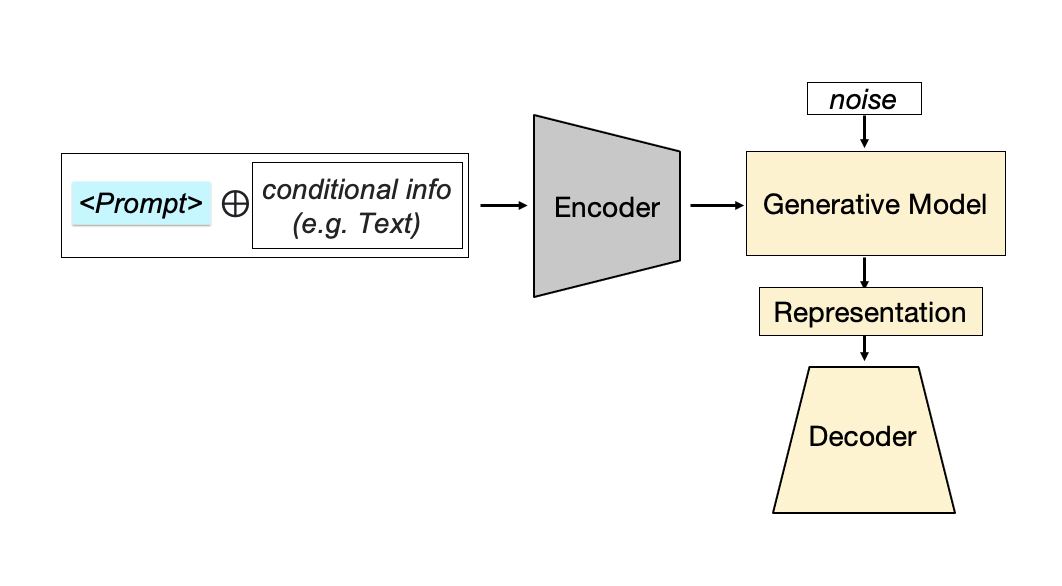} \vspace{-0.5cm}
    \caption{Prompting Generative Models. A depiction of a typical text-to-image generation framework, detailing elements such as conditional information, an image encoder, a generative model, noise injection, latent space representation, and a decoder. Conditional information can take various forms such as hard prompts, learnable soft prompts, or a combination of the two. Furthermore, prompts can be presented in textual, visual, or both formats.}
    \label{fig:gen} 
\end{figure}

This section provides an overview of the preliminaries required to understand the prompting model in text-image generation, with a specific focus on diffusion models. 

Text-image generation automatically synthesis vivid and realistic images from natural language descriptions and has attracted much more attention. From the pioneering work DRAW~\cite{gregor2015draw}, text-image generation models have seen numerous breakthroughs. Generative adversarial network (GAN)~\cite{goodfellow2020generative} then used to design end-to-end differentiable image generation structure~\cite{reed2016generative} which is followed by many works~\cite{pan2023drag, karras2019style, isola2017image}. Besides, variational auto-encoder (VAE)~\cite{kingma2019introduction} is also adapted to generate images~\cite{pu2016variational, vahdat2020nvae}. However, these models are trained on small-scale data and lack generalization~\cite{ramesh2021zero}. Autoregressive methods driven by large-scale datasets, such as DALL-E~\cite{ramesh2021zero}, and Parti~\cite{yu2022scaling}, are proposed and demonstrate surprising zero-shot generation ability. Recently, the diffusion model (DM) has spurred another line of state-of-the-art models for text-image generation~\cite{dhariwal2021diffusion}. 
Diffusion models, also known as diffusion probabilistic models~\cite{sohl2015deep},  originate from non-equilibrium statistical physics~\cite{jeulin1997dead} and sequential Monte Carlo~\cite{neal2001annealed} and are designed to fit any data distribution while keeping tractable. The denoising diffusion probabilistic models (DDPMs)~\cite{ho2020denoising} first adopt DMs in the image generation domain and inspire the whole community of generative models. In inference, DDPMs build a Markov chain that generates images from noisy data within finite transitions which is called \textit{reverse process}. In training, DDPMs learn from the \textit{forward process} where noise is added to the natural images and estimated by the model. Given a clean image $x_0$ from a distribution $q$, diffusion step $T$ and hyperparameters $\beta_t$, the forward process generates $x_T$ following
\begin{equation}
    q(x_{1:T}|x_0) \coloneqq \prod_{t=1}^T q\left(x_t \mid x_{t-1}\right),
\end{equation}

\begin{equation}
    q\left(x_t \mid x_{t-1}\right)\coloneqq\mathcal{N}\left(x_t ; \sqrt{1-\beta_t} x_{t-1}, \beta_t I\right).
\end{equation}

The noised image from any arbitrary step $t$ then can be reformulated as 
\begin{equation}
    q\left(x_t \mid x_0\right)\coloneqq\mathcal{N}\left(x_t ; \sqrt{\bar{\alpha}_t} x_0,\left(1-\bar{\alpha}_t\right) I\right),
\end{equation}
where $\alpha_t\coloneqq 1-\beta_t, \bar{\alpha_t}\coloneqq \prod^t_{s=0}\alpha_s$.

Given the defined forward process, the DDPM is trained in the reverse process which starts from $p_{\theta}(x_T)$ by the loss defined as 

\begin{equation}
    L(\theta)\coloneqq\mathbb{E}_{t, \mathbf{x}_0, \boldsymbol{\epsilon}}\left[\left\|\boldsymbol{\epsilon}-\boldsymbol{\epsilon}_\theta\left(\sqrt{\bar{\alpha}_t} \mathbf{x}_0+\sqrt{1-\bar{\alpha}_t} \boldsymbol{\epsilon}, t\right)\right\|^2\right],
\end{equation}
where $t$ is uniform between $1$ and $T$, $\boldsymbol{\epsilon} \sim \mathcal{N}(\mathbf{0}, \mathbf{I})$ is random noise, and $\boldsymbol{\epsilon}_{\theta}$ is known as noise predictor parametrized by $\theta$. 

By incorporating additional control information, typically in the form of textual prompts, the efficacy of the reverse process in diffusion models has been significantly enhanced to control the synthesis results rather than random sampling. This textual-based generation has solidified its position as the pioneering foundation in the field of text-to-image generation. Consequently, let $\Gamma$ be an encoder that maps a conditioning input prompt $P$ into a conditioning vector $\boldsymbol{c}\coloneqq \Gamma(P)$, the conditioned learning objective has been expanded with $\boldsymbol{c}$ that represents textual prompt. 
\begin{equation}
    L(\theta)\coloneqq\mathbb{E}_{t, \mathbf{x}_0, \boldsymbol{\epsilon}, \boldsymbol{c}}\left[\left\|\boldsymbol{\epsilon}-\boldsymbol{\epsilon}_\theta\left(\sqrt{\bar{\alpha}_t} \mathbf{x}_0+\sqrt{1-\bar{\alpha}_t} \boldsymbol{\epsilon}, t, \boldsymbol{c}\right)\right\|^2\right],
\end{equation}

Figure 2 illustrates a typical text-image generation framework, highlighting its key components and functionalities, including (1) fixed or learnable conditional information, such as hard textual prompts or learnable soft prompts. The conditional information can be in textual form or in other modalities; (2) An encoder $\mathcal{E}$ of the input image; (3) A generative model, such as diffusion model, autoregressive model, or GAN; (4) Noise injection or interference; (5) A representation of features in the latent space or low-resolution images; (6) A decoder $\mathcal{D}$ for image decoding or super-resolution for a high-fidelity generation. The training process involves dataset utilization, loss functions, and optimization techniques to train the model for generating coherent and visually appealing images based on text prompts. During the inference stage, the trained model is utilized to generate images based on user-specified prompts. The formulation of prompts plays a crucial role as it governs communication with the model and influences the desired outcomes of image generation. This section focuses on prompt engineering in text-image generation and its applications.

\subsection{Understanding Prompting} \label{sec 5.2}
To gain a deeper understanding of the factors that influence the generated images, we will introduce prompt design in text-to-image from the view of semantics, prompt diversity, and controllable prompts.

\noindent\textbf{Semantic Prompt Design. }
The art of prompt semantics has significant impacts on image generation in diffusion models~\cite{witteveen2022investigating}. 
The linguistic components such as adjectives, nouns, and proper nouns in prompt influence image generation in different ways but consistently. While descriptors (simple adjectives) subtly affect the output, nouns introduce new content more effectively. Interestingly, using an artist's name tends to generate images deviating significantly from the original, and incorporating lighting phrases can dramatically modify image content and mood. Therefore, the quality of image generation can be enhanced through clear, noun-based statements, effective seeds, and the emulation of artist styles.

\noindent\textbf{Diversify Generation with Prompt.}
Apart from direct handcrafting individual prompts in a semantic way, recent works experiment with various prompt modifiers $\mathcal M$ focusing on enhancing the diversity of initial prompts $P$ by $\Tilde{P} = \mathcal M(P)$ with $\Tilde{P}$ be the diversified prompts. 
DiffuMask~\cite{wu2023diffumask} explores two strategies in prompt modifiers $\mathcal M$, \ie, retrieval-based prompt and prompt with Sub-class, with $P$ set to \textit{"Photo of a [sub-class] car in the street"}. Specifically, they retrieve real images and captions sets~\cite{beaumont2022clip, radford2021learning}, with captions as the prompt sets for generating synthetic images. Besides, they select sub-classes from Wiki based on the main class. 
ImaginaryNet~\cite{ni2022imaginarynet} uses GPT2~\cite{radford2019language} as $\mathcal M$ with a given class name $y$ of the target object to generate a complete description of an imaginary scene $\Tilde{P}_y$ under the guidance of prefix phrases of \textit{"A photo of a"}. The prompt serves as generating diversified photo-realistic imaginary images for the imaginary supervised object detection task. 
Similarly,~\cite{he2023synthetic} uses a word-to-sentence T5 model~\cite{raffel2020exploring} as $\mathcal M$ to generate detailed prompts $\Tilde{P}_y$ targeted for a specific label space $y$, thereby maximizing the potential of synthesized data in data-scarce settings by enriching the diversity of prompts.
These approaches further obtain diversified images $I$ by $I = \mathcal{G}(\boldsymbol{\epsilon}|\Tilde{P})$ where $\mathcal{G}$ represents the generative model.

\noindent\textbf{Complex Control of Synthesis Results.}
As the synthesized image generation is usually inconsistent due to noise injection and randomness lying in the stochastic nature of diffusion models, recent work has been emerging in the area of complexly controllable generation. To avoid controllability limitations with user-provided masks that restrict the modified area~\cite{avrahami2022blended, nichol2021glide}, prompt-based control is gaining attention. OneWord~\cite{gal2022image} aims to solve the problem of generating personalized images with specific subjects that are hard to describe with pure texts. Therefore they proposed a prompt method that designates a placeholder string $S_{\ast}$ to represent the new concept such as \textit{"a photograph of $S_{\ast}$ on the beach"} with its associated learned embedding $v_\ast$.
A similar design is done by DreamBooth~\cite{ruiz2023dreambooth}. Instead of creating new words, 
they design prompts with \textit{(unique identifier, subject)} pairs that bind rare tokens from T5-XXL tokenizer~\cite{raffel2020exploring} as unique identifiers for the specific subjects and the coarse class name of the subjects, such as \textit{"A [V] dog"}, with \textit{[V]} as the rare-token identifiers. They further retain the representation of class names in prompts by introducing extended class-prior preservation loss to the training objective. 
Custom Diffusion~\cite{kumari2022multi} extended the customization into a multi-concept scenario where multiple personalized concepts are composed in the same generated image, such as family members in the same family photo. They design prompts at this aim by including the use of a unique modifier token \textit{$S^*_i$} for each concept \textit{$i$}, initialized with different rare tokens and positioned ahead of category namex.

Over the days, textual prompts only cannot meet the specific needs of image-processing tasks, and controllable text-to-image generation is gaining attention~\cite{feng2022training, epstein2023diffusion, kawar2023imagic}. A wide range of task-specific input conditions, such as canning edge encoded by image encoder~\cite{canny1986computational}, are added with trainable network architecture to the diffusion model in the work of ControlNet~\cite{zhang2023adding}. The additional task-specific conditions $\boldsymbol{c}_{\mathrm{f}}$ is added to the overall training objective as 
$\left.L(\theta)\coloneqq\mathbb{E}_{t, \boldsymbol{x}_0, \boldsymbol{\epsilon}, \boldsymbol{c}, \boldsymbol{c}_{\mathrm{f}}}\left[\| \boldsymbol{\epsilon}-\boldsymbol{\epsilon}_\theta\left(\boldsymbol{x}_t, t, \boldsymbol{c}, \boldsymbol{c}_\mathrm{f}\right)\right) \|_2^2\right]$.
Notably, in order to improve the semantic recognition ability of the encoder from control maps and optimize ControlNet's performance even when explicit prompts are absent, ControlNet's training utilizes a method where half of the text prompts are randomly replaced with empty strings.

Controlling the synthesis results can also be done after the generation process with prompt editing methods. To bypass the common demand of user-defined spatially fixed masks~\cite{avrahami2022blended, nichol2021glide}, Prompt-to-Prompt~\cite{hertz2022prompt} can edit images by only editing prompts by replacing a word, specifying a style, changing adjectives, \etc. The manipulations are infiltrated by injecting the cross-attention maps controlling which pixels attend to which tokens of the prompt text during which diffusion steps. Prompt-based image editing methods that merely modify the text prompt provide more intuitive editing experiences.

\subsection{Application of Prompting}
Text-to-image diffusion models, aided by prompting techniques, have excelled in data generation applications. This section investigates their efficacy in generating training data that boost the scope and flexibility of learning procedures. Additionally, we explore the versatility of these models in crafting diverse data in target domains, spanning diverse output formats like images, videos, 3D models, and motion. Also, we unveil its potential in complex task-solving and adversarial attacks.

\subsubsection{Generating Synthetic Training Data} Recent advancements have sparked a growing interest in prompting text-to-image models as innovative synthesized training data generators for various tasks downstream tasks such as segmentation, object detection, and image recognition. Challenges such as data scarcity and the need for high-resolution synthetic images can be mitigated through intricate prompt engineering. DiffuMask~\cite{wu2023diffumask} automatically generates high-resolution synthetic training images with the aforementioned prompt engineering strategies in Sec.~\ref{sec 5.2}. Its created pixel-level semantic masks between prompts and generated images can be seamlessly applied for segmentation tasks, including semantic segmentation, open-vocabulary segmentation, and domain generalization on real images.
ImaginaryNet~\cite{ni2022imaginarynet} generates synthesis data to tackle the challenge of insufficient real images and annotations for training object detection. It generates scene descriptions with LLM from class labels and prompts the text-to-image model for creating imaginary training data.
Under different training settings of pure or mixed imaginary and real data, object detectors are enhanced for the Imaginary Supervised Object Detection task (ISOD), especially under settings where real images and annotations are unavailable.
Synthetic data is also proven feasible for image recognition tasks, specifically in zero-shot and few-shot settings. ~\cite{he2023synthetic} creates the synthetic data for image recognition in a two-phase manner. Firstly, novel samples are synthesized using target category names. Secondly, a fine-tuned language model is used to convert category names into richly contextual, diversified language prompts for diversifying the training data.

\subsubsection{Generating Data in Target Domain} In addition to the role of training data generators, diffusion models also play a pivotal role as target data generators. Importantly, their capabilities extend beyond the generation of images. They can efficiently generate video data, three-dimensional data, and motion data, further broadening their application range and utility.

\noindent\textbf{Text-to-Video Generation.}
Make-A-Video from~\cite{singer2022make} is the first approach for directly translating the tremendous recent progress in text-to-image (T2I) generation to text-to-video (T2V) without paired text-video data. It infers actions and events in the prompt and generates video by leveraging joint text-image priors to bypass the need for paired text-video data. 
Imagen Video~\cite{ho2022imagen} propels T2V generation towards a more efficient stage, delivering higher video resolution outputs by combining a frozen T5 text encoder~\cite{raffel2020exploring}, a base video diffusion model, and interleaved spatial and temporal super-resolution diffusion models, \ie, cascaded diffusion models~\cite{ho2022cascaded}.
However, works on T2V commonly face challenges in editing capabilities and effective training on specific domains. FateZero~\cite{qi2023fatezero} overcomes these limitations with a zero-shot text-based editing method capable of editing attributes, style, and shape on real-world videos without per-prompt training or use-specific mask. Specifically, FateZero utilizes a pair of user-provided source prompt $P_{src}$ and the editing prompt $P_{edit}$. The source prompt is for obtaining a noisy latent representation $\boldsymbol{x}_t$ of the source video frame, then $\boldsymbol{x}_t$ is denoised conditioned on the editing prompt $P_{edit}$.
Tune-A-Video~\cite{wu2022tune} tackles the challenge of computational expensiveness with the one-shot tuning strategy on one text-video pair and only on the first and former video frames. This study is in the inspiration that T2I models attend well to verbs in the prompt in generating still images and exhibit surprisingly good motion consistency alignment with prompts when extended to T2V. Tune-A-Video is also equipped with editing capability by capturing essential motion information from the input video and synthesizing novel videos with edited prompts preserving the motion words. Moreover, textual prompt-based generation has advanced to multi-modal generation, \eg generating simultaneously aligned audio-video pairs~\cite{ruan2023mm, zhu2023moviefactory}.

\noindent\textbf{Text-to-3D Generation.}
Previous works face challenges of insufficient large-scale labeled 3D datasets and inefficient architectures for denoising 3D data. As a consequence, prompt-based generation has advanced from T2I to T2V models and also in text-to-3D scenarios where high-quality 3D objects and scenes are generated from text prompts~\cite{muller2023diffrf}. 
DreamFusion~\cite{poole2022dreamfusion} firstly randomly initializes the 3D object with NeRF~\cite{mildenhall2021nerf}  for each text prompt and produces 2D image renderings $x=g(\eta)$ with differentiable image generator $g(\eta)$. These renderings are generated from various angles and paired with view-dependent prompts prefixes such as \textit{"overhead view"} and \textit{"front view"} and then diffused and reconstructed by Imagen~\cite{ho2022imagen} with $q\left(x_t \mid x_0\right)\coloneqq q\left(g(\eta)_t \mid g(\eta)_0\right)$. The sampled noise $\boldsymbol{\epsilon}$ guides a gradient direction to be backpropagated to the NeRF parameters $\eta$.
To tackle the issue in the growing popular DreamFusion regarding the optimization efficiency of NeRF which leads to low-quality 3D models with a long processing time, Magic3D~\cite{lin2022magic3d} proposed a two-phase coarse-to-fine optimization framework, \ie, firstly obtaining coarse diffusion prior from text prompts with Imagen~\cite{ho2022imagen} and then rendering efficiently with high-resolution latent diffusion models (LDM)~\cite{rombach2022high}. Borrowing the idea from~\cite{ruiz2023dreambooth}, Magic3D is capable of personalized prompt-based editing of 3D models by binding the \textit{[V]} identifier in the prompt with the 3D object.
Besides, prompt-based editing can be done through finetuning with LDM in the coarse-to-fine stage with the modified prompt. 
Inaccurate and unfaithful structures in text-to-3D generation due to random shape initialization without prior knowledge lead Dream3D~\cite{xu2022dream3d} to explicit 3D shape priors into the CLIP-guided 3D optimization process~\cite{poole2022dreamfusion, lee2022understanding, khalid2022clip}. Specifically, it connects the T2I model and a shape generator as the text-to-shape stage to produce a 3D shape prior with shape components in the prompts. Then it harnesses the 3D shape prior to the initialization of NeRF and optimizes it with the full prompt. To close the gap between the synthesis image and shape, and also inspired by ~\cite{gal2022image, ruiz2023dreambooth}, Dream3D links renderings with stylized text prompt suffixes in the format of \textit{"a CLS in the style of $\ast$"} where \textit{CLS} represents the shape category and \textit{$\ast$} is a placeholder token that requires optimization of its text embedding jointly with the weights of Stable Diffusion for capturing the style of the rendered images.

\noindent\textbf{Text-to-Motion Generation.}
Another area where the power of prompt-based generation is exemplified is in the realm of text-to-motion (T2M).
MotionDiffuse~\cite{zhang2022motiondiffuse} is a diffusion model-based text-driven motion generation framework with motion sequence as the input $\textbf{x}_0$. It has a body part-independent controlling scheme that generates separate sequences for each body part under $m$ fine-grained prompts $P_i$ with $i\in [1,m]$ for each body part $i$ and predicts each $\epsilon_i^{part} =\epsilon_\theta\left(\mathbf{x}_t, t, \Gamma(P_i)\right)$. Besides, it generates arbitrary-length continuous motion synthesis using time-varied text prompts with $m$ intervals, denoted as array $\left\{P_{i, j},\left[l_{i, j}, r_{i, j}\right]\right\}$ and predicts the $\epsilon_i^{time}$. All noises are interpolated mutually with other parts for the continuous motion sequence generation.
Similarly, as in T2I, T2V, and text-to-3D, it is also required for T2M synthesis with flexible editing capability.  
Thus, FLAME~\cite{kim2022flame} enables editing with free-form language description with novel transformer-based diffusion architecture. It takes diffusion time-step tokens, motion length tokens, language tokens, and motion tokens as input tokens to the transformer and can therefore handle motion sequences of variable length. MDM~\cite{tevet2022human} also introduces editability and controllability with a similar idea borrowed from image inpainting by adding suffixes and prefixes to the motion in the temporal domain. And the textual condition guides MDM to fill the missing body part with a specific motion while keeping the rest intact in the spatial domain.

\noindent\textbf{Complex Conditional Scene Generation.} The use of diffusion models has expanded beyond single target data generation, finding applications in various scenarios that involve generating more complex scenes tailored to specific use cases with more complex conditional inputs. In robotics, text guidance is used to perform aggressive data augmentation on top of our existing robotic manipulation datasets to generate robotic scenes via inpainting various unseen objects for manipulation, backgrounds, and distractors~\cite{yu2023scaling}. In autonomous driving, diffusion models are leveraged to generate controllable pedestrian trajectories that align with the surrounding environment's context that enables the simulation of realistic pedestrian behavior~\cite{rempeluo2023tracepace}. Additionally, diffusion models can incorporate conditional information in the form of graphs that represent individual rooms to generate house floorplans, facilitating the design and planning of residential spaces~\cite{shabani2023housediffusion}.

\subsubsection{Prompt-centered Complex Task} 
Beyond the former direct applications of text-to-other generation, prompt-centered complex application in various scenarios reveals the field's true versatility and potential. In the context of storytelling, StoryBook~\cite{jeong2023zero} retains a visual narrative storybook with consistent character faces through a series of prompt-centered steps. It first generates prompts of scene descriptions with LLM, which are prompted to the latent diffusion model with designated special token placeholder $S_\ast$ like~\cite{gal2022image}, to ground consistent character faces during generation.
Similarly,~\cite{lu2023multimodal} proposed multimodal procedure planning (MPP) task, where the initial stepwise textual plan is generated with LLM and then serves as prompts to diffusion model for synthesizing text-grounded image plan. What's different is that the image plans are verbalized through image captioning backward to the LLM for revising the initial plan showing the potential for multimodal prompting.

\subsection{Responsible AI Considerations of Prompting}
\label{sec:t2i-ethical}
Artificial Intelligence is revolutionizing our world through its formidable learning ability, transformative force, and profound influence across diverse areas of society. It also spurred intense debate about ethical issues, principles, and integrity in AI development and applications. There is a global convergence around five ethical principles~\cite{jobin2019global}: transparency, justice and fairness, non-maleficence, responsibility, and privacy. In this subsection, we discuss ethical issues when prompting text-to-image generative models.  

\noindent
\textbf{Adversarial Robustness of Prompt.}
The adversarial attacks have been introduced to text-to-image diffusion models for mainly 2 aims. Some work takes diffusion models as a tool to facilitate or defend against adversarial attacks~\cite{chen2023diffusion, nie2022diffusion}. Some work directly attacks diffusion models~\cite{zhuang2023pilot} and aims to erase image content given character perturbations. As the pioneer to introduce diffusion models in the adversarial attack field, DiffAttack~\cite{chen2023diffusion} unveils the potential of diffusion models for crafting adversarial examples with satisfactory imperceptibility and transferability by manipulating the latent space rather than pixel space. This approach maintains visual quality with embedding perturbations undetectable to humans and transferable across diverse model architectures. Diffusion models can be utilized for adversarial purification - a defense strategy that removes adversarial perturbations. DiffPure~\cite{nie2022diffusion} implements this approach, adding a minimal amount of noise to an adversarial example before reversing the generative process to restore the original image, thus exhibiting robust defense capabilities against powerful adaptive attacks. Zhuang \etal~\cite{zhuang2023pilot} study the query-free attack generation on Stable Diffusions where an adversarial text prompt is obtained in the absence of end-to-end model queries. They show the vulnerability of Stable Diffusions rooted in the text encoders. A five-character text perturbation is able to shift the output content. 

\noindent
\textbf{Backdoor Attack of Prompt Learning.}
Backdoor attacks on text-to-image generative models aim to control the content of generated images during inference by embedding inputs with predefined backdoor triggers. The attacker secretly injects backdoors, such as specific text characters, into the model during training to trigger the model to either generate images with pre-defined attributes or images following a hidden or even malicious description. The backdoor attack may lead to inappropriate outputs such as offensive content. On the other hand, it can also be used in copyright protection by watermarking the models. Struppek \etal~\cite{struppek2022rickrolling} demonstrate that the text encoders pose a major tampering risk. The attack is a teacher-student approach and only involves fine-tuning a text encoder by generating backdoor targets and triggers on the fly. Zhai \etal~\cite{zhai2023text} design three types of backdoor attacks, namely pixel-backdoor, object-backdoor, and style-backdoor, and demonstrate the text-to-image diffusion models' vulnerability to backdoor attacks. 
Huang \etal~\cite{huang2023zero} explore the vulnerability to backdoor attacks via personalization for a more efficient attack. Text-to-image personalization guides the diffusion-based text-to-image model to generate user-provided novel concepts through natural language. Huang \etal~\cite{huang2023zero} devised backdoor attacks on two families of personalization methods, Textual Inversion~\cite{gal2022image} and DreamBooth~\cite{ruiz2023dreambooth}. 

\noindent
\textbf{Fairness and Bias.}
Generative AI models are typically trained on web-scale datasets scraped from the internet and are inevitable to biased human behavior as shown in~\cite{friedrich2023fair, naik2023social, wang2023T2IAT, luccioni2023stable}. For example, Stable Diffusion only generates images with white male-appearing persons as firefighters~\cite{friedrich2023fair}. Some studies start to pay more attention to the fairness issues related to text-to-image generations and can be grouped into three paradigms: 1) training data pre-processing to remove bias before learning~\cite{smith2023balancing, seth2023dear}, 2) enforcing fairness during training by introducing constraints on the learning objective~\cite{seth2023dear}, 3) post-processing approaches to modify the model outcome at the deployment stage~\cite{friedrich2023fair, chuang2023debiasing, kim2023explaining}. 

\noindent
\textbf{Privacy.}
There might be privacy-sensitive information, \eg face identity, in the huge amount of training data for training text-to-image models. Such information may arise privacy risks in real-world applications such as information leaks. Membership inference attacks are an approach to investigating privacy leakage by inferring whether a specific data sample was used in the training phase (called member or non-member respectively)~\cite{wu2022membership}.  Some work~\cite{wu2022membership, duan2023diffusion, webster2023reproducible} studies the privacy risks of text-to-image generation models from the perspective of membership attacks. From the perspective of prompting, Shen \etal~\cite{shen2023prompt} propose \textit{prompt stealing attack}, which steals prompts from images generated by text-to-image generation models. The creation of high-quality prompts can be challenging, time-consuming, and costly. Hence successful prompt stealing attacks direct violate intellectual property and even jeopardize the business model of prompt trading markets.

\section{Prompting VLM vs. Uni-modal Models}
\label{sec:unimodal}
\subsection{Prompting in Natural Language Processing}
This section summarizes existing studies on prompt engineering on textual language models. Prompt engineering has been widely adopted in various natural language processing applications including question answering~\cite{khashabi-etal-2020-unifiedqa, jiang2021can}, text classification~\cite{gao2021making, lester-etal-2021-power}, text generation~\cite{radford2019language, brown2020language, schick-schutze-2021-shot}, and information extraction~\cite{KnowPrompt, cui-etal-2021-template}, \etc. Recent LLMs such as InstructGPT~\cite{ouyang2022training} and PALM2~\cite{anil2023palm} have shown incredible generalized inference ability through prompting. Early works~\cite{paranjape2021prompting} designed natural language templates to let pre-trained language models fill in to explain their predictions. Wei \etal~\cite{wei2022chain} demonstrate that the performance of LLMs can be significantly improved by adding intermediate reasoning steps into the prompt. In particular, the prompt of each task contains a few manual demonstrations consisting of a question and a reasoning chain leading to the answer. The LLM learns to follow the prompt and thinks step-by-step to solve the given task. Liu \etal~\cite{liu2021makes} find that the quality of the prompt, \ie, the selection of examples in prompts and given explanations, largely impacts LLM's performance. Fu \etal~\cite{fu2022complexity} demonstrate that prompting LLMs with complex example questions, which requires more intermediate reasoning steps, could achieve better performance and benefit the model's robustness regarding format perturbation and distribution shift. 

Manually crafting prompts for each task strongly depends on human experience, and manual testing would be required to evaluate and improve the template, which would be time-consuming. Zhang \etal~\cite{zhang2022automatic} work on eliminating manual efforts by leveraging LLMs to construct reasoning chains with demonstrations. Besides,  a line of works~\citep{lewis2020retrieval, borgeaud2022improving, izacard2022few} automates the prompt engineering by utilizing a dense retriever to augment the language models with external resources, which has also been referred to as retrieval-augmented language models. For a given question, the dense retriever retrieves relevant text from a knowledge source and appends it to the language model input. Such language models have recently demonstrated strong performance on knowledge-intensive tasks. 
\cite{lester2021power} propose \textit{prompt tuning} that appends the input embedding layer with extra trainable tokens and learns these tokens through backpropagation on downstream tasks. This opens a direction of learning soft prompts to enhance LLMs.

Many studies demonstrated that LLMs' performance considerably drops as the task complexity increases. A natural way for humans to solve complex tasks is to decompose them into a series of simple subtasks and solve the complex task by completing each simple subtask. A line of works investigated enhancing LLMs' performance on complex tasks by prompting LLMs multi-times, where the LLMs are expected to solve a subtask by each prompt. Press \etal~\cite{press2022measuring} examine the capacity of language models to execute compositional reasoning tasks and found that LLMs are good at memorizing facts but do not compose them to answer questions. To narrow the compositionality gap, the authors let LLMs ask themselves follow-up questions, answer the questions, and decide whether they have sufficient information to give the final solution. Kazemi \etal~\cite{kazemi2022lambada} propose a backward chaining algorithm to decompose a complex task by starting from the objective and recursively breaking down the complex task into sub-tasks based on rules. Khot \etal~\cite{khot2022decomposed} decompose a complex task into sub-tasks and use sub-task-specific LLMs to solve them, leading to improved performance on a line of textual multi-step reasoning tasks.

Researchers have also noticed ethical and integrity issues related to prompt engineering in NLP. Yang \etal~\cite{yang2022prompting} propose a prompt-based adversarial attack to compromise NLP models and robustness enhancement techniques. This work indicates that the prompting paradigm has the potential in probing fundamental vulnerabilities of large language models and fine-tuning them for downstream tasks. Dong \etal~\cite{dong2023promptattack} adopt a prompt-based learning approach to automatically generate effective adversarial examples to probe Dialogue State Tracker models. The prompt may inherit the bias in the pre-trained models and~\cite{delobelle-etal-2022-measuring} review the literature on fairness metrics for pre-trained language models and experimentally evaluate compatibility. Moreover, one can refer to several existing surveys~\cite{Lou2023Is, Ding2022OpenPrompt, qiao2022reasoning} for a more comprehensive review.

\subsection{Prompting on Pure Vision Models}
Although prompt is first widely adopted in natural language models, many works also utilize prompts in pure vision models~\cite{kirillov2023segment, bahng2022exploring, wang2022images, wang2023seggpt, loedeman2023prompt, tu2022visual, zhang2022promptcal} and applications including image classification~\cite{bahng2022exploring, loedeman2023prompt, tu2022visual, zhang2022promptcal}, image segmentation~\cite{kirillov2023segment, wang2022images}, depth estimation~\cite{wang2022images}, keypoint detection~\cite{wang2022images}, denoising~\cite{wang2022images}, detaining~\cite{wang2022images}, and image enhancement~\cite{wang2022images} \etc.

 Several studies have identified two main mechanisms for incorporating prompts into vision models. The first mechanism treats prompting as an adaptation method that facilitates the fine-tuning of pre-trained vision models~\cite{bahng2022exploring, salman2021unadversarial, loedeman2023prompt}. The second mechanism utilizes prompts as a module that plays a role in both model pre-training and inference~\cite{kirillov2023segment, wang2022images, zhang2022promptcal}.

 Pre-trained vision models have significantly improved performance, but their size has also increased drastically, making training and fine-tuning infeasible for most users. To address this issue, adapting pre-trained models to specific tasks in a parameter-efficient way is critical. Many studies have treated prompting as an adaptation method.

 Bahng \etal~\cite{bahng2022exploring} use a single visual prompt to adapt a frozen large-scale vision model to a new task. Adaptation approaches such as fine-tuning and linear probes require some level of access to the pre-trained model during both training and testing. However, visual prompting only requires model access during training, making it feasible for some applications~\cite{salman2021unadversarial}. Additionally, Tu \etal~\cite{tu2022visual} propose Visual Query Tuning (VQT) to adapt pre-trained Transformers to downstream tasks while keeping the backbone frozen, allowing for more accurate predictions utilizing the intermediate features of a pre-trained model.

 As classical fine-tuning methods become more limiting when models are hosted as inference APIs, visual prompt learning is emerging as a potential solution for adapting frozen and cloud-hosted models. Loedeman \etal~\cite{loedeman2023prompt} introduce the Prompt Generation Network (PGN), which generates input-dependent visual prompts to facilitate domain adaptation. PGN generates new prompts for every image by combining items from a commonly learned library of tokens. It consists of a lightweight neural network that learns the probability distribution for selecting prompt vectors from a token library.

 In addition to using prompts as an adaptation for downstream tasks, some researchers have integrated prompting modules into the entire model to improve pre-training performance and enable more flexible inference. In~\cite{kirillov2023segment}, Kirillov \etal~ introduce the Segment Anything Model (SAM), which aims to build a foundation model for segmentation. Inspired by prompting techniques in NLP, they proposed the \textit{promptable segmentation task} to generate a valid segmentation mask based on any segmentation prompt. The prompt can include spatial or text information that identifies an object in the image, and the output of the corresponding model is a reasonable mask for at least one target object. This promptable segmentation task is used in both pre-training and downstream segmentation tasks.

 Painter, presented in~\cite{wang2022images}, is a generalist model that can perform various vision tasks based on given task prompts. It can perform tasks such as semantic segmentation, instance segmentation, depth estimation, keypoint detection, denoising, detailing, and image enhancement, as well as out-of-domain tasks like open-category object segmentation. To address the issue of general-purpose prompt definition, Painter formulates the dense-prediction vision problem as \textit{image inpainting}. This way, input/output paired images from the same task can be used as input to indicate the task the model should perform.

 Following the work on Painter, Wang \etal propose SegGPT~\cite{wang2023seggpt}, which focuses on the segmentation task and enables segmentation of everything with a generalist Painter. Zhang \etal utilize auxiliary prompts to approach the Generalized Novel Category Discovery (GNCD) setting by proposing a prompt-based Contrastive Affinity Learning (PromptCAL) method~\cite{zhang2022promptcal}. Existing semi-supervised learning methods fail to learn unlabeled data from novel semantic classes, but PromptCAL is discriminative to novel semantic information.

 The combination of prompt engineering with visual models has also triggered a line of work focusing on integrity and ethics issues. Chen \etal~\cite{chen2023visual} leveraged visual prompting to improve the adversarial robustness of a fixed, pre-trained model at testing time. Li \etal~\cite{li2023exploring} explored the benefits of visual prompting in constructing compelling neural network classifiers with differential privacy. However, such studies are still relatively rare and more attention is required. 

\section{Challenges and Opportunities}
\label{sec:7-discussion}

\textbf{Prompting Model in Multimodal-to-Text Generation.}
In addition to visual and textual modalities, the incorporation of other modalities such as audio and thermal is possible. It is crucial to address the inherent heterogeneity among these modalities, which includes variations in data formats, scales, and structures.

Two notable projects in this domain are Kosmos~\cite{huang2023language, peng2023kosmos}, developed by Microsoft, and IMAGEBIND~\cite{Girdhar_2023_CVPR}, developed by Meta AI. These projects aim to create unified models capable of handling diverse modalities, promoting the utilization of such unified models as a significant direction in the field.

However, it is important to note that most of the research on prompts for multimodal-to-text pre-trained models has primarily focused on hard prompts. Conversely, soft prompts based on image-text matching models, such as CLIP~\cite{radford2021learning}, have received extensive investigation. Models like CoOp~\cite{zhou2022learning} and CoCoOp~\cite{zhou2022conditional} leverage soft prompts on it to enhance model performance. Nevertheless, the exploration of prompt tuning for popular generative multimodal-to-text pre-trained models remains largely unexplored.

Additionally, multimodal-to-text pre-trained models employ a range of challenging prompt techniques, including in-context learning~\cite{alayrac2022flamingo} and instruction tuning~\cite{li2023otter}. Despite their effectiveness, the underlying mechanisms by which these models learn and the specific contributions of different aspects of the demonstrations remain largely unexplored. A deeper understanding of these factors is crucial for refining and optimizing the performance of multimodal-to-text pre-trained models. 

\noindent\textbf{Prompting Model in Image-Text Matching.}
Although pre-trained encoders prompted by a matching loss have been widely used for adaptation in downstream tasks, the exploration of visual prompting on pre-trained encoders remains relatively unexplored. Similar to the seamless adaptation of textual encoders through learning textual prompts, the investigation of visual prompts is an intriguing area that can unlock emergent abilities, especially in difficult scenarios such as dense objects, object hallucination, and the adaptation to modern VLMs. In the future, it is imperative to address questions regarding the specific type of visual prompts that are essential and the semantic information that these prompts introduce. By delving into these inquiries, we can gain a deeper understanding of the role and impact of visual prompts, thereby further advancing the field.

Meanwhile, the investigation of how unified prompting can enhance the performance of both branches remains understudied. In an intuitive sense, a unified prompt can provide us with referential information that spans across modalities, as discussed in~\cite{yao2022cpt}. This has the potential to facilitate the development of multimodal models that are capable of visual grounding and enable referential dialogues encompassing visual and textual co-reference.

\noindent\textbf{Prompting Model in Text-Image Generation.}
One of the significant challenges in the field of prompting text-to-other generation models, particularly in the case of Text-to-Video (T2V) and Text-to-3D (T2-3D) models, is their dependency on Text-to-Image (T2I) models. These models often share the same concern due to the nature that they are extensions of T2I diffusion models. For example, the inconsistency of the input control maps in T2I models can lead to errors in the consequently generated videos and 3D objects, thereby affecting the overall performance and reliability of these extension scenarios.

Looking ahead, there are several promising directions for future research. One such direction is the incorporation of visual prompting into T2I, T2V, and T2-3D diffusion models. In the context of text-to-image generation, visual prompting and visual annotations can offer more visual cues, leading to the creation of more personalized images. This approach allows for more attention to be paid to specific areas of the image, enhancing the detail and accuracy of the generated output.
Visual prompts can also be beneficial for video generation, either on a frame-by-frame basis or for T2-3D generation aimed at improving 2D renderings or shapes. The concept of visual prompts can be further expanded to include video prompts, object prompts, and motion prompts, depending on the specific requirements of different target data generation scenarios. 
Furthermore, text-image matching models hold the potential for better alignment as multi-modal prompting in the generation. This approach could lead to more accurate and contextually relevant image generation, opening up new possibilities for the application of pre-trained vision-language models.

\noindent\textbf{Generalizing Prompting Methods from Unimodal to Multimodal.} 
Sec.~\ref{sec:unimodal} discusses the applications of prompt engineering in both pure vision and pure language models, which can motivate further research in multi-modality research. When combined with instruction-tuning methods, pure language models have enabled phenomenal applications such as ChatGPT~\cite{chatgpt}. The potential of these methods such as Reinforcement Learning from Human Feedback (RLHF)~\cite{ouyang2022training} and Harmlessness from AI Feedback~\cite{bai2022constitutional} can be further explored in multimodal models as shown in several recent studies~\cite{li2023otter, gao2023llamaadapterv2}. Constitutional AI is a method proposed in~\cite{bai2022constitutional} to train a harmless AI assistant through self-improvement, without any human labels identifying harmful outputs. Although some efforts have been put into language models~\cite{bai2022constitutional}, how to implement constitutional AI in the multimodal domain is still an open question. 

Another potential direction is the adoption of in-context prompts in multimodal models. Large unimodal language models can address a specific new task given several demonstrations of the task in their text prompt without any gradient update. Flamingo~\cite{alayrac2022flamingo} has also demonstrated the few-shot in-context learning ability, but how to further improve the in-context learning capacity is still under-explored.

\noindent\textbf{Responsible AI Considerations of Prompting.}
There are already a few studies concerning ethical issues on multimodal-to-text generation in Sec.~\ref{sec:m2t-ethical} and text-to-image generations as discussed in Sec.~\ref{sec:t2i-ethical}.  Integrity and ethical issues of prompt engineering on vision-language models need much more attention. One possible direction is to prevent bias and backdoor attacks inherited from the pre-trained model during downstream prompt adaptations~\cite{carlini2022poisoning,gao2023backdoor,huang2022backdoor,yang2023backdoor}. As large models are normally pre-trained on web-scale datasets which may preserve biased knowledge or sensitive privacy information, the post-deployment procedure conducted by prompt engineering should be able to control the potential risks.

Adversarial Robustness has been intensively studied in various model architectures, such as Convolutional Networks~\cite{madry2017towards,jia2022adversarial}, Vision Transformers~\cite{wu2022towards,gu2022vision}, and Capsule Networks~\cite{gu2021effective}. It has not been fully understood how the prompting on VLMs with both a vision architecture and a language component performs under adversarial attacks. Especially, the impact of recent advances in VLM on adversarial robustness remains to be studied. E.g., do large prompts bring robustness to VLMs~\cite{gu2023towards}?

Besides, transparency and controllable generation through fair prompting are also essential in generative tasks. Generative models are shown to be vulnerable to privacy leakage~\cite{wu2022membership, duan2023diffusion, webster2023reproducible} and may generate biased content ~\cite{friedrich2023fair, naik2023social, wang2023T2IAT, luccioni2023stable}. Hence, constructing transparent and controllable prompts that are capable to conserve privacy and prevent unethical generation is critical for real-world applications. Last but not least, managing the accompanying risks of prompt engineering and large models requires the close collaboration of society, research institutions, and government~\cite{anderljung2023frontier, hacker2023regulating}.

\noindent\textbf{Relationship between Prompts on Different VLMs.} The recent work~\cite{li2022dall} studies the relationship between concepts learned by multimodal-to-text and image-to-text and text-to-image models. They show the studied two types of models cannot fully understand each other, while they also share some concepts. Similarly, the relationship between prompts on different types of models should be explored in future work, especially the feasibility of building universal prompts across different models trained on the same data. In addition to the inter-model relationship, the interaction between prompts and model architecture should be investigated since most prompts are proposed on Transformer-based models. Concretely, how the model's self-attention changes during prompting~\cite{gu2022vision}.

\section{Conclusion}

This survey paper on prompt engineering of pre-trained vision-language models has provided valuable insights into the current state of research in this field. The main findings and trends identified through the analysis shed light on the effective utilization of prompts in adapting large pre-trained models for vision-language tasks.

One key finding is the versatility and applicability of prompt engineering across different types of vision-language models, including multimodal-to-text generation models, image-text-matching models, and text-to-image generation models. The survey explored each model type from their respective characteristics, highlighting various prompting methods on them.

The implications of these findings are significant for both academia and industry. By leveraging prompt engineering techniques, researchers can achieve remarkable performance gains in vision-language models without the need for extensive labeled data. This has the potential to reduce the burden of data annotation and accelerate the deployment of vision-language models in real-world applications.

However, it is important to acknowledge the limitations of this survey. The rapidly evolving nature of the field and the wide range of existing prompt engineering approaches make it challenging to provide an exhaustive overview. Additionally, the survey focused primarily on pre-trained vision-language models from a prompting engineering perspective and may not have covered all recent advancements in other related areas.

To address these limitations, we will maintain and release a platform to keep tracking the advance in this area. Further research should explore the integration of prompt engineering techniques with other emerging technologies, such as reinforcement learning or meta-learning, to enhance the performance and generalization capabilities of vision-language models. Additionally, investigations into the interpretability and robustness of prompt-engineered models are crucial for ensuring their practical deployment and ethical use.

Overall, this survey contributes to the existing body of knowledge by providing a comprehensive overview of prompt engineering in pre-trained vision-language models. By elucidating the current state, key trends, and implications of prompt engineering techniques, this survey serves as a valuable resource for researchers and practitioners aiming to harness the potential of vision-language models for various applications. It fills a gap in research by offering insights into the adaptation of pre-trained models in the context of vision and language, paving the way for further advancements in this exciting field.

\section*{Acknowledgements}
We would like to thank Ananth Balashankar (Google Research) and Ashkan Khakzar (University of Oxford) for constructive feedback on an earlier version of this manuscript.

\renewcommand\refname{References}

{\small
\bibliographystyle{unsrt2authabbrvpp}
\bibliography{ref}
}

\end{document}